\pdfoutput=1

\documentclass[11pt]{article}

\usepackage[]{acl}

\usepackage{times}
\usepackage{latexsym}

\usepackage[T1]{fontenc}

\usepackage[utf8]{inputenc}

\usepackage{microtype}

\usepackage{inconsolata}

\usepackage{graphicx} 
\usepackage{natbib}  
\usepackage{caption} 
\usepackage{multirow}
\usepackage{algorithm}
\usepackage{algpseudocode}
\usepackage{amsmath}
\usepackage{booktabs}
\usepackage{amsfonts}
\usepackage{blkarray, bigstrut}
\parskip=\medskipamount
\usepackage{amssymb}
\usepackage{pifont}
\newcommand{\cmark}{\ding{51}}%
\newcommand{\xmark}{\ding{55}}%

\usepackage[labelfont=bf]{caption}
\usepackage{multirow}

\usepackage{newfloat}
\usepackage{listings}

\usepackage{xcolor}
\usepackage[capitalise,noabbrev]{cleveref}

\newcommand{\nonumberfootnote}[1]{
    \begingroup
    \let\thefootnote\relax
    \footnotetext{#1}
    \endgroup
}

\title{Product Description and QA Assisted Self-Supervised Opinion Summarization}

\author{Tejpalsingh Siledar$^*$$^\clubsuit$,
Rupasai Rangaraju$^*$$^\clubsuit$,  Sri Raghava Muddu$^*$$^\clubsuit$, 
\\ 
\textbf{Swaprava Nath$^\clubsuit$, Pushpak Bhattacharyya$^\clubsuit$,} 
\\
\textbf{Suman Banerjee$^\spadesuit$, Amey Patil$^\spadesuit$, Sudhanshu Shekhar Singh$^\spadesuit$,}\\
\textbf{Muthusamy Chelliah$^\spadesuit$, Nikesh Garera$^\spadesuit$}\\
        $^\clubsuit$Computer Science and Engineering, IIT Bombay, India, \\
        $^\spadesuit$Flipkart, India \\
        \texttt{\{tejpalsingh, rupasai, sriraghava, swaprava,
        pb\}@cse.iitb.ac.in}
        \\
        }

\begin{document}
\maketitle

\begin{abstract}
  In e-commerce, opinion summarization is\nonumberfootnote{*\,Equal contribution.} the process of summarizing the consensus opinions 
  found in product reviews. However, the potential of additional sources such as \textit{product description} and \textit{question-answers (QA)} has been considered less often. Moreover, the absence of any supervised training data makes this task challenging. 
  To address this, we propose a novel synthetic dataset creation (SDC) strategy that leverages information from reviews as well as additional sources for selecting one of the reviews as a pseudo-summary to enable supervised training. Our \underline{M}ulti-\underline{E}ncoder \underline{D}ecoder framework for \underline{O}pinion \underline{S}ummarization (\textbf{MEDOS}) employs a separate encoder for each source, enabling effective selection of information while generating the summary. For evaluation, due to the unavailability of test sets with additional sources, we extend the Amazon, Oposum+, and Flipkart test sets and leverage ChatGPT\footnote{\url{https://chat.openai.com/} (gpt-3.5 August 3 version)} to annotate summaries. Experiments across nine test sets demonstrate that the combination of our SDC approach and MEDOS model achieves on average a $\mathbf{14.5\%}$ improvement in ROUGE-1 F1  over the SOTA.  Moreover, comparative analysis underlines the significance of incorporating additional sources for generating more informative summaries. Human evaluations further indicate that MEDOS scores relatively higher in \textit{coherence} and \textit{fluency} with $0.41$ and $0.5$ ($-1$ to $1$) respectively, compared to existing models. \textit{To the best of our knowledge}, we are the first to generate opinion summaries leveraging additional sources in a self-supervised setting.
\end{abstract}
\section{Introduction}
In the e-commerce domain, reviews play a vital role in making informed decisions. However, due to the recent proliferation of online reviews, going through all the product reviews before making a decision is challenging. Opinion summarization provides a solution by summarizing the opinions presented in the reviews \citep{hu2006opinion, wang-ling-2016-neural, angelidis-lapata-2018-summarizing}. 
\begin{table}[t]
    \centering
    \resizebox{\columnwidth}{!}{%
    \begin{tabular}{p{\columnwidth}}
    \toprule
    \small{\textbf{MultimodalSum}}\\
    \midrule
    \small{I bought this product to scan my negatives. It does not work with Windows XP. I have tried to contact the company several times and have not received a response. I am very disappointed in the product. I would not recommend it to anyone.}\\
    \midrule
    \small{\textbf{Our Model (MEDOS)}}\\
    \midrule
    \small{I purchased the \textbf{VuPoint FS-C1-VP Film and Slide Digital Converter} to scan my \textbf{35mm} film and slide negatives. It is not compatible with Windows XP. \underline{The} \underline{software does not work with Windows 7 or 8}. I have tried to contact the company and they do not respond to my emails. I would not recommend this product to anyone.}\\
    \bottomrule
    \end{tabular}%
    }
    \caption{MultimodalSum vs. MEDOS generated summary for a product from the Amazon test set. Information assisted from product description and question-answers are in \textbf{bold} and \underline{underline} respectively. Our model is able to capture essential information from the product description and question-answers, not found in reviews. This makes our model-generated summaries more informative while still retaining the consensus opinions from reviews as evident in the above example.}
    \label{Table: example}
\end{table}
However, text summarization \citep{Nallapati2016AbstractiveTS, See2017GetTT, Liu2019TextSW} usually contains reference summaries which are very difficult to obtain at a large scale for opinion summarization. As a result, recent studies \citep{brazinskas-etal-2020-unsupervised,elsahar-etal-2021-self} enable self-supervision by curating synthetic pairs out of review corpus by sampling one of the reviews as a pseudo summary and considering the remaining reviews as the input. 

\noindent \textbf{\underline{Motivation}}\; In e-commerce, users' opinions are expressed through various sources such as product ratings, reviews, review upvotes and downvotes, and question-answers. Additionally, for each product, description, product specification, product images, price, etc. are present as well. Considering such additional sources apart from reviews is vital in generating opinion summaries that are well-rounded and informative.
Specifically, descriptions offer nuanced details about various aspects, while question-answers provide additional perspectives on specific queries, both of which can be valuable. 
Table \ref{Table: example} shows an example of the influence of product description and question-answers. 
However, acquiring annotated training datasets proves expensive and impractical as the number of sources increases.
This makes it essential to devise effective synthetic dataset creation strategies that enable supervised training of models using multiple sources.


\noindent \textbf{\underline{Problem Statement}}\; We propose a novel synthetic dataset creation approach that uses additional sources such as \textit{product description} and \textit{question-answers (QA)} along with reviews for generating synthetic quadruplets of the form \textit{\{input reviews, description, question-answers, pseudo-summary\}} to enable end-to-end supervised training. A \underline{m}ulti-\underline{e}ncoder \underline{d}ecoder model for \underline{o}pinion \underline{s}ummarization (\textbf{MEDOS})  to effectively select information from either product description or question-answers while summarizing reviews. For evaluation, due to the unavailability of test sets that have annotated summaries written considering such additional sources (except for Flipkart \citep{Siledar2023AspectSentimentbasedOS}), we extend the available e-commerce test sets by including these additional sources and leveraging ChatGPT \citep{openai2023} to annotate \citep{Gilardi_2023, Huang_2023} summaries. \\
\noindent\textbf{Input}: \textit{Reviews, Description, Question-Answers}\\
\textbf{Output}: \textit{Opinion Summary}

\noindent Our contributions are:
\begin{enumerate}
    \item A novel synthetic dataset creation (SDC) approach that enables supervised training in the presence of additional sources without the need for any annotated training datasets. We propose a \underline{M}ulti-\underline{E}ncoder \underline{D}ecoder framework for \underline{O}pinion \underline{S}ummarization (\textbf{MEDOS})\footnote{Code and data: \url{https://github.com/tjsiledar/MEDOS}} to effectively fuse information from \textit{reviews, product description,} and \textit{question-answers (QA)} 
    (Section \ref{problem}, \ref{sdc} \& \ref{model_framework}). \textit{To the best of our knowledge}, we are the first to do multi-source self-supervised opinion summarization.
    \item Extensions to e-commerce test sets namely Amazon \citep{brazinskas-etal-2020-unsupervised} and Oposum+ \citep{amplayo-etal-2021-aspect} to include additional sources. For comparison, we extend: \textbf{Amazon}, \textbf{Oposum+}, and \textbf{Flipkart} by curating six new test sets: Amazon R, Amazon RDQ, Oposum+ R, Oposum+ RDQ, Flipkart R, and Flipkart RDQ leveraging ChatGPT to annotate summaries. We extend the test sets to contain $\mathbf{662}$ \textbf{opinion summaries} across six curated test sets (Section \ref{tdc}, Table \ref{Table: data}).
    \item Experimental demonstrations of our SDC approach and MEDOS model in outperforming the SOTA model on \textbf{nine test sets} on average by $\mathbf{14.5\%}$ in ROUGE-1 F1 (Section \ref{results}).
    \item Comparative and qualitative analysis indicating the importance of sources such as \textit{product description} and \textit{question-answers} in generating more informative summaries compared to existing models (Section \ref{results}, Table \ref{Table: dummy} \& \ref{Table: example_summaries}).
\end{enumerate}
%
\begin{table*}
    \centering
    \resizebox{2\columnwidth}{!}{%
    \begin{tabular}{lcccccccccc}
    \toprule
    &\multicolumn{3}{c}{\textbf{Original}} &\multicolumn{6}{c}{\textbf{Extended (Ours)}}\\
    \cmidrule(lr){2-4}\cmidrule(lr){5-10}
    \textbf{} & Amazon & Oposum+ & Flipkart & Amazon & Oposum+ & Flipkart & Amazon & Oposum+ & Flipkart\\
    & &&& GPT-R & GPT-R & GPT-R & GPT-RDQ & GPT-RDQ & GPT-RDQ\\
    \midrule
    \#domains & $4$ & $6$ & $3$ & $4$ & $6$ & $3$ & $4$ & $6$ & $3$ \\
    \#test set & $32$ & $30$ & $145$ & $32$ & $30$ & $145$ & $32$ & $30$ & $145$\\
    \#reviews/product & $8$ & $10$ & $10$ & $8$ & $10$ & $10$ & $8$ & $10$ & $10$\\
    \#summaries/product & $3$ & $3$ & $1$ & {$\mathbf{3}$} & {$\mathbf{3}$} & {$\mathbf{1}$}  & {$\mathbf{3}$} & {$\mathbf{3}$} & {$\mathbf{1}$}\\
    \#summaries & $96$ & \underline{$90$} & $145$ & {$\mathbf{96}$} & {$\mathbf{90}$} & {$\mathbf{145}$} & {$\mathbf{96}$} & {$\mathbf{90}$} & {$\mathbf{145}$}\\
    \#descriptions & - & - & - & - & - &- & $21$ & $17$ & $145$ \\
    \#question-answers & - & - & - & - & - &- & $11$ & $10$ & $145$ \\
    \bottomrule
    \end{tabular}%
    }
    \caption{\textbf{Statistics for original and extended test sets.} GPT-R indicates the use of \textit{reviews} whereas GPT-RDQ indicates the use of \textit{reviews}, \textit{description}, and \textit{question-answers} to generate summaries using ChatGPT. \textbf{Bold} represents our contributions. In the respective extended versions, reviews are the same as the original.}
    \label{Table: data}
\end{table*}

\section{Related Work}\label{related_appendix}
\textbf{Self-supervised Opinion Summarization.}\; 
Recent approaches use self-supervision by considering one of the reviews as a pseudo-summary. \citet{brazinskas-etal-2020-unsupervised}
randomly selected $N$ reviews per entity to construct $N$ pseudo-summary, reviews pairs. \citet{amplayo-lapata-2020-unsupervised} sampled a review randomly and generated noisy versions of it as input reviews. \citet{Amplayo2020UnsupervisedOS}  used aspect and sentiment distributions to sample pseudo-summaries. \citet{elsahar-etal-2021-self} selected reviews similar to a randomly sampled pseudo-summary as input reviews, based on TF-IDF cosine similarity.  \citet{wang-wan-2021-transsum} aimed at reducing opinion redundancy and constructed highly relevant reviews pseudo-summary pairs by learning aspect and sentiment embeddings to generate relevant pairs. \citet{im-etal-2021-self} used synthetic dataset creation strategy similar to \citet{brazinskas-etal-2020-unsupervised} and extended it to multimodal version. \citet{Ke2022ConsistSumUO} captured the consistency of aspects and sentiment between reviews and pseudo-summary using constrained sampling. \citet{siledar-etal-2023-synthesize} use lexical and semantic similarities for creating synthetic datasets. Our work is most similar to \citet{elsahar-etal-2021-self} in using cosine similarity to select input reviews and pseudo-summary pairs. However, we use review embeddings 
to compute similarity instead of TF-IDF scores. Additionally, our pseudo-summary selection considers additional sources such as \textit{product description} and \textit{question-answers} as well. Our synthetic dataset creation strategy ensures that the pseudo-summary selection is highly relevant to all our input sources. Recent opinion summarization systems \citep{bhaskar-etal-2023-prompted, hosking-etal-2023-attributable} include a large number of reviews. However, we limit our work to a fixed number of reviews to enable a fair comparison with previous approaches.

\noindent\textbf{Additional sources for Opinion Summarization.}\; \citet{zhao2020weakly} used aspects identified from product description to perform extractive aspect-based opinion summarization. \citet{Li_Yuan_Xu_Wu_He_Zhou_2020} proposed a supervised multimodal summarization model to effectively generate summaries using reviews, product image, product title, and product details. \citet{im-etal-2021-self}
proposed a self-supervised multimodal training pipeline to generate summaries using reviews, images, and meta-data. \citet{Siledar2023AspectSentimentbasedOS} did supervised opinion summarization using simple rules to generate summaries separately in the form of verdict, pros, cons, and additional information using reviews, description, specifications, and question-answers. Our work takes inspiration from \citet{im-etal-2021-self} to utilize a multi-encoder framework to effectively fuse information from various sources. However, where additional sources are all text, our approach of forming highly relevant synthetic pairs using additional sources helps in capturing relevant information. Also, our approach differs from \citet{Siledar2023AspectSentimentbasedOS} in training models in an end-to-end fashion without the aid of supervised summaries. 

\section{Problem Formulation}\label{problem}
\textbf{Preliminaries.} For a specific product or an entity, $R = \{r_1,...,r_N\}$ is the set of $N$ reviews, 
$D$ represents the product description, 
and $Q = \{q_1,...,q_M\}$ represents a set of $M$ question-answer pairs such that $q_i$ represents the $i^{th}$ concatenated question and its corresponding answer.

\noindent\textbf{Opinion Summarization.} The task of opinion summarization is to generate an opinion summary $s$ given a set of reviews $R$ for an entity (eg. product or business). \citet{rush-etal-2015-neural} defined the task of abstractive summarization as:
\begin{align}
    s^* &= \underset{s}{\text{argmax}}\; g(s, R), \\
    g(s, R) &= \log \; p(s|R;\theta),\\
    &\approx \sum_{i=0}^{J-1} \log \; p(s_{i+1}| s_w, R; \theta),
\end{align}
where $g$ is a scoring function defined as a conditional log probability of the summary given the input, $s_w = s_{[i-w+1,...,i]}$ for a window size $w$, $\theta$ is the neural network parameters, and $|s| = J$. For opinion summarization, the input is a review set $R$ and the output is the opinion summary $s$.
The conditional probability can be modeled using Transformers \citep{NIPS2017_3f5ee243} as:
\begin{align}
    p(s_{i+1}| s_w, R; \theta) &\propto \nonumber
    \\ 
    \rho&(\text{FFN}(\text{C-Attn}(\mathbf{a_{R}}, \mathbf{e_{s_w}}))), \\
    \mathbf{a_R} = \text{S-Attn}(\text{Enc}&(R)), \;
    \mathbf{e_{s_w}} = \text{Emb}(s_w),
\end{align}
where $\rho$ is the softmax function, FFN is the feed-forward network, C-Attn is the cross-attention network, S-Attn is the self-attention network, Enc is the encoder, and Emb is the embedding layer.

\begin{figure}[]
    \centering
    \includegraphics[width=0.8\columnwidth]{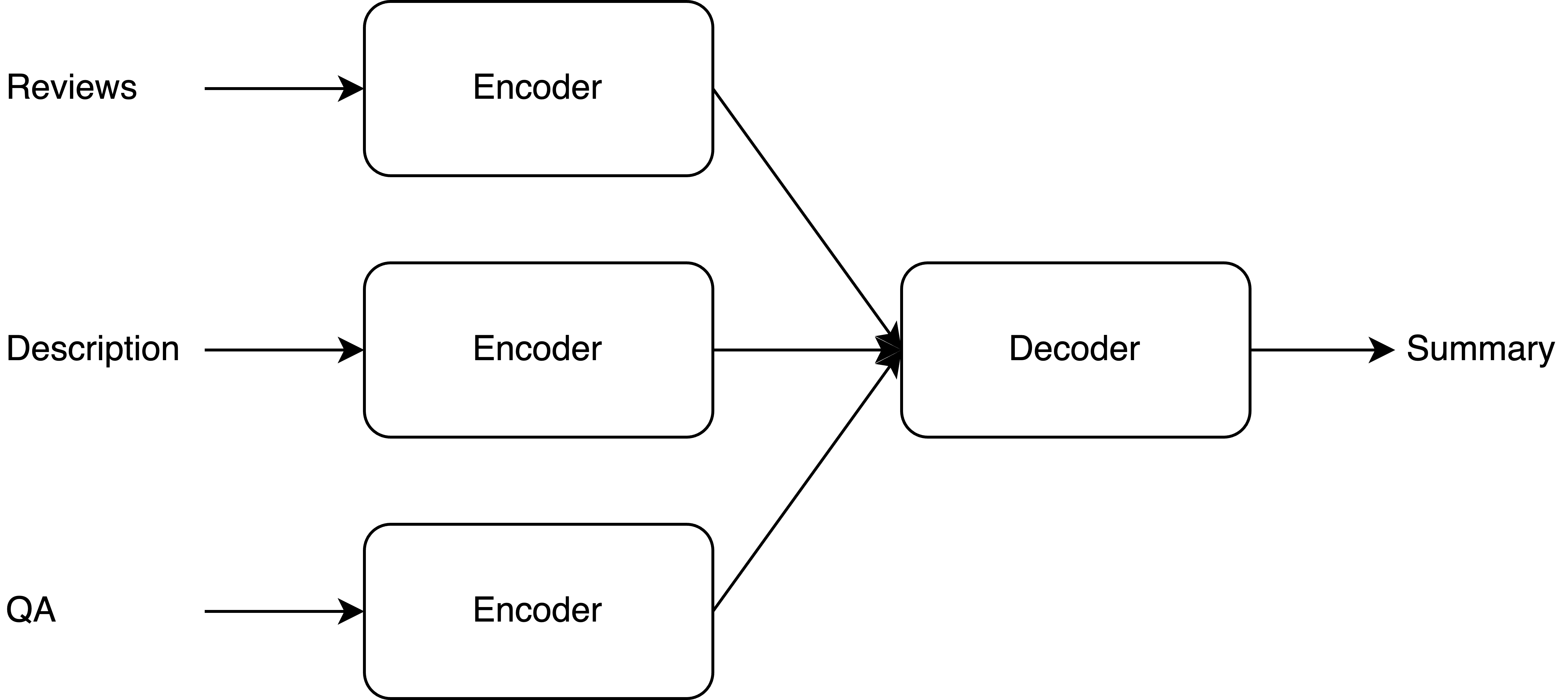}
    \caption{Framework of our MEDOS model that takes \textit{reviews}, \textit{description}, and \textit{question-answers (QA)} as the input. During inference, the model generates a summary whereas during training the model uses pseudo-summary obtained through SDC process for learning.} 
    \label{fig:model_arch}
\end{figure}

\noindent\textbf{Additional Sources.} Under the presence of additional sources such as product description and question-answers, the equations for modeling abstractive summarization can be written as:
\begin{align}
    s^* &= \underset{s}{\text{argmax}}\; g(s, R, D, Q), \\
    g(s, R, D, Q) &= \log \; p(s|R, D, Q;\theta),\\
    \approx \sum_{i=0}^{J-1} &\log \; p(s_{i+1}| s_w, R, D, Q; \theta),
\end{align}
Using transformers, this can be modeled as:
\begin{align}
    p(s_{i+1}| s_w, R,D,Q; \theta) &\propto \nonumber \\
    \rho(\text{FFN}&(\text{C-Attn}(\mathbf{a_{f}}, \mathbf{e_{s_w}}))), \\
    \mathbf{e_{s_w}} = \text{Emb}(s_w),
\end{align}
where $\mathbf{a_f}$ is the fused attention. We propose a Multi-Encoder Decoder Framework- MEDOS (Section \ref{model_framework}, Figure \ref{fig:model_arch}) to create fused attention $\mathbf{a_f}$ (Eq. \ref{eq11}).

\begin{table*}[h]
    \centering
    \small
    \resizebox{2\columnwidth}{!}{%
    \begin{tabular}{clcccccccccccc}
    \toprule
    &&&&& \multicolumn{3}{c}{\textbf{Amazon}} & \multicolumn{3}{c}{\textbf{Amazon GPT-R}} & \multicolumn{3}{c}{\textbf{Amazon GPT-RDQ}} \\
    \cmidrule(lr){6-8}\cmidrule(lr){9-11}\cmidrule(lr){12-14}
    \textit{abs?} &\textbf{Model} &\textbf{R}& \textbf{D}& \textbf{Q} & R1 $\uparrow$ & R2 $\uparrow$ & RL $\uparrow$ & R1 $\uparrow$ & R2 $\uparrow$ & RL $\uparrow$ & R1 $\uparrow$ & R2 $\uparrow$ & RL $\uparrow$ \\
    \midrule
         \xmark &Random  &\cmark & \xmark & \xmark  & $27.86$ & $3.87$ & $16.68$   & $20.69$ & $1.56$ & $12.55$  & $18.83$ & $1.45$ & $12.03$   \\
         \xmark &Oracle  &\cmark & \xmark & \xmark  & $44.47$ & $13.83$  & $30.85$  & $33.69$ & $6.04$ & $22.88$  & $31.83$ & $5.77$ & $22.04$   \\
     \midrule
    \xmark  & Clustroid  &\cmark & \cmark & \cmark & $29.27$ & $4.41$ & $17.78$   & $22.74$ & $2.16$ & $14.03$  & $21.31$ & $2.57$ & $13.38$   \\
    \xmark &LexRank    &\cmark & \cmark & \cmark     & $29.46$ & $5.53$ & $17.74$ & $22.82$ & $3.08$ & $13.77$ & $19.30$ & $4.31$ & $12.90$   \\
    \xmark &QT  &\cmark & \cmark & \cmark & $34.04$ & $7.03$ & $18.08$     & $23.01$ & $2.48$ & $12.05$ & $21.78$ & $3.25$ & $12.36$    \\
    \midrule
    \cmark &CopyCat    &\cmark & \xmark & \xmark     & $31.97$ & $5.81$ & $20.16$  & $20.09$ & $1.79$ & $12.94$ & $20.54$ & $1.94$ & $13.85$    \\
    \cmark &PlanSum  &\cmark & \xmark & \xmark & $32.87$ & $6.12$ & $19.05$  & $20.49$ & $1.76$ & $12.44$  & $19.09$ & $1.58$ & $12.02$      \\
    \cmark &ConsistSum   &\cmark & \xmark & \xmark   & $33.32$ & $5.94$ & {$\mathbf{21.41}$}  & - & - & - & - & - & -  \\
    \cmark &MultimodalSum &\cmark & \cmark & \xmark & $34.19$ & $7.05$ & $20.81$ & $21.43$ & $1.58$ & $13.20$ & $20.39$ & $2.08$ & $12.83$  \\
    \cmark &TransSum &\cmark & \xmark & \xmark  & $34.23$ & \underline{$7.24$} & $20.49$   & - & - & -  & - & - & - \\
    \cmark &COOP &\cmark & \xmark & \xmark  & {$\mathbf{36.57}$} & $7.23$ & \underline{$21.24$}   & - & - & -  & - & - & - \\
    
    \midrule
            \cmark & T$5$-concat &\cmark & \cmark & \cmark  & $28.04$ & $4.46$  & 
    $16.39$  &   $21.28$ & {$\mathbf{2.57}$} & $13.00$ & $20.61$ & $2.72$ & $13.33$ \\
    \cmark & BART-concat &\cmark & \cmark & \cmark  & $32.35$ & $6.49$  & 
    $19.78$  &   \underline{$22.32$} & \underline{$2.27$} & \underline{$13.74$} & \underline{$21.75$} & \underline{$2.39$} & \underline{$13.57$} \\

    \midrule
    \cmark & \textbf{MEDOS} &\cmark & \cmark & \cmark  & \underline{$34.63$}  & {$\mathbf{7.48}$}  & 
    $20.97$   &   {$\mathbf{23.92}$*} & \underline{$2.27$*} & {$\mathbf{14.69}$*} & {$\mathbf{25.44}$*} & {$\mathbf{4.16}$*} & {$\mathbf{16.45}$*} \\
    \bottomrule
    \end{tabular}%
    }
    \caption{\textbf{Results on Amazon test set and its extensions.} \textbf{R}, \textbf{D}, \textbf{Q} indicate the presence of \textit{reviews}, \textit{description}, and \textit{question-answers} respectively in the input. \textit{abs?} indicate abstractive systems. \textbf{Bold} and \underline{underline} indicate best and second-best scores using abstractive systems. * indicates pvalue $<0.05$ on paired t-test against MultimodalSum. Overall our combination of SDC approach and MEDOS outperforms baselines across all three test sets.}
    \label{Table: amazon_results}
\end{table*}%

\begin{algorithm}[t]
\caption{SDC using Additional Sources}\label{alg:cap}
\begin{algorithmic}[1]
\Require Reviews $R$, $\mathbf{e_R} \in \mathbb{R}^{N \times d}$, product description $D$, $\mathbf{e_D} \in \mathbb{R}^{1 \times d}$, and question-answer pairs $Q$, $q \in Q$, $\mathbf{e_q} \in \mathbb{R}^{1 \times d}$ for a product. Functions $sim$, $diag$, and $mean$.
\\
\textbf{Initialize} $Z = []$
\For{each product}
    \State $M \gets diag(sim(\mathbf{e_R},\mathbf{e_R}), 0)$ 
    \hfill $\footnotesize{\{\in \mathbb{R}^{N \times N}\}}$
    \State $ds \gets sim(\mathbf{e_R},\mathbf{e_D})$ \hfill $\{\in \mathbb{R}^{N \times 1}\}$
    \For{$q \in Q$}
        \State $qs  \mathrel{+}= sim(\mathbf{e_R},\mathbf{e_q})$ \hfill$\{\in \mathbb{R}^{N \times 1}\}$
    \EndFor
    \State $qs \gets mean(qs)$ \hfill$\{\in \mathbb{R}^{N \times 1}\}$
    \State $ss \gets \lambda_{1} \cdot ds + \lambda_{2} \cdot qs$ 
    \State $R_p \gets \text{top-p reviews using}\; ss$
    \For{$r \in R_p$}
        \State $T \gets \text{top-k reviews for $r$ using}\; M$ 
        \State $Z.insert(\{T, D, Q, r\})$
    \EndFor
\EndFor
\\\textbf{Return} $Z$
\end{algorithmic}
\end{algorithm}
\section{Synthetic Dataset Creation (SDC)}\label{sdc}
Before discussing the details of our framework, we formalize the synthetic dataset creation process used to train these models. In the absence of supervised datasets, most recent approaches \citep{brazinskas-etal-2020-unsupervised, im-etal-2021-self} resort to self-supervision wherein \{input reviews, pseudo-summary\} pairs are constructed. 

Following \citet{brazinskas-etal-2020-unsupervised}, we can assume that a review $r \in R$ can serve as a summary for a set of reviews $T \subseteq R-\{r\}$. This lets us create training points $(T,r)$ i.e. \{input reviews, pseudo-summary\}, similar to what the model will experience during inference. $T$ is fixed to size $k$, 
enabling comparison with existing works.

However, in the presence of additional sources such as product description $D$ and question-answer pairs $Q$, we slightly modify this definition. Instead of synthetic pairs, we construct synthetic quadruplets of the form: \{input reviews, product description, question-answers, pseudo-summary\}.

Algorithm \ref{alg:cap} details the process of generating synthetic quadruplets. We generate multiple such quadruplets out of reviews $R$, product description $D$, and question-answer pairs $Q$ for a specific product. The overall idea for synthetic dataset creation is to choose relevant quadruplets for training. Here we define relevance as the quadruplet that best aids our model in learning the task of opinion summarization using multiple sources.

The intuition is to first select a pseudo-summary $r$ that is the closest to both $D$ and $Q$. We measure closeness in terms of cosine similarity $sim$ between their embeddings (SBERT \citep{reimers-2019-sentence-bert}).  This selection ensures that the pseudo-summary $r$ contains information relevant to both $D$ and $Q$ so that the model learns to pick information from these two sources as well during training. Next, using the pseudo-summary $r$ selected, we look for its closest $k$ set of reviews that can act as its input reviews set $T$, which ensures that the model learns the task of summarization.

More formally, we first compute a matrix $M \in \mathbb{R}^{N \times N}$ by computing cosine similarity between embeddings of each review pair $(r_a, r_b)$ where $r_a,r_b \in R$. We make all the diagonals of $M$ as zero to remove self-comparisons using $diag$ function. Next, we compute $ds \in \mathbb{R}^{N \times 1}$ by computing cosine similarity between the embeddings of each review $r_a$ and $D$. We also compute $qs \in \mathbb{R}^{N \times 1}$ by computing cosine similarity between the embeddings of each review $r_a$ and all $q \in Q$ and taking a $mean$ of it respectively. Finally, we compute $ss \in \mathbb{R}^{N \times 1}$ as $\lambda_{1} \cdot ds + \lambda_{2} \cdot qs$ where $\lambda_1, \lambda_2$ are parameters set to $0.5$ for our experiments. We select $R_p \subseteq R$ reviews for forming $p$ synthetic quadruplets by taking the top-p scores from $ss$. For each review $r \in R_p$, we get the top-k reviews $T$ from $R-\{r\}$ using scores corresponding to the review $r$ from $M$. This lets us form synthetic quadruplet instances such as $\{T, D, Q, r\}$ for model training.

\section{Model Framework (MEDOS)}\label{model_framework}
Figure \ref{fig:model_arch} represents our multi-encoder framework, where each source passes through its separate encoder to generate separate attentions: $\mathbf{a_R} = \text{S-Attn}(\text{Enc}(T)),\;\mathbf{a_D} = \text{S-Attn}(\text{Enc}(D)),$ and $\;\mathbf{a_Q} = \text{S-Attn}(\text{Enc}(Q))$. The fused attention $\mathbf{a_f}$ is then computed as:
\begin{align}
    \mathbf{a_f} &= \mathbf{a_R} + \alpha \odot \mathbf{a_D} + \beta \odot \mathbf{a_Q} \label{eq11}
\end{align}
where $\odot$ represents element-wise multiplication, $\alpha$ and $\beta$ act as gates regulating the flow of information from product description and question-answers, computed as: $\alpha = \phi([\mathbf{a_R};\mathbf{a_D}]\mathbf{W_\alpha})$ and $\beta = \phi([\mathbf{a_R};\mathbf{a_Q}]\mathbf{W_\beta})$ where $\mathbf{W_\alpha, W_\beta}$ are learned parameters and $\phi(\mathbf{x}) = \text{\small RELU}(\text{\small tanh}(\mathbf{x}))$ is the activation, following \citet{im-etal-2021-self}. 

\begin{table*}
    \centering
    \resizebox{2\columnwidth}{!}{%
    \begin{tabular}{p{0.5\columnwidth} | p{0.5\columnwidth} | p{0.5\columnwidth} | p{0.5\columnwidth}}
    \toprule
        \small{\textbf{Product $\mathbf{1}$}} & \small{\textbf{Product $\mathbf{2}$}} & \small{\textbf{Product $\mathbf{3}$}} &
        \small{\textbf{Product $\mathbf{4}$}}\\
    \midrule
    \small{I purchased the \textbf{VuPoint FS-C1-VP Film and Slide Digital Converter} to scan my \textbf{35mm film} and slide negatives. It is not compatible with Windows XP. \underline{The software does not work} \underline{with Windows 7 or 8}. I have tried to contact the company and they do not respond to my emails. I would not recommend this product to anyone.}
    &
    \small{The \textbf{Marpac TSC 330 Travel Sound Conditioner} is a great little machine. It is small enough to travel with, but big enough to be used at home. The sound quality is great and it is easy to use. The only thing I don't like about it is that it doesn't have a volume control.} 
    &
    \small{The \textbf{Sony Speaker Dock} is a great product. The sound is great and the remote control works great. The only thing I don't like about it is that it doesn't charge my iphone 4s. I have to \underline{buy an adaptor} for that.}
    &
    \small{The \textbf{Opteka HG-1 Heavy-Duty Aluminum Ultra HandGrip Handheld Stabilization System for DSLR and Video Cameras} is a great product. I use it with my \underline{Nikon Coolpix L820} and it works great. It is a \underline{little heavy}, but that is to be expected for a small camera.}
    \\
        \bottomrule
    \end{tabular}%
    }
    \caption{\textbf{Qualitative Analysis.} MEDOS generated summaries for four different products from the Amazon test set utilizing \textit{reviews, description, and question-answers}. Information assisted by the product description is indicated in \textbf{bold}, whereas those assisted from the question-answers are \underline{underlined}.}
    \label{Table: dummy}
\end{table*}

\section{Experiments}
\subsection{Datasets}
We conducted experiments on: Amazon \citep{amazondata,brazinskas-etal-2020-unsupervised}, Oposum+ \citep{amplayo-etal-2021-aspect}, and Flipkart \citep{Siledar2023AspectSentimentbasedOS}. Statistics are in Table \ref{Table: data}. Using our SDC strategy, we created $387$k and $313$k instances from the Amazon and
Oposum+ respectively to enable supervised training. Due to the unavailability of review data in the case of Flipkart, we used the Amazon data to train models. Refer \textbf{Appendix \ref{dataset_appendix}}.



\subsection{Test Dataset Extension}\label{tdc}
In the absence of any test sets that contain additional sources, we extended Amazon, Oposum+, and Flipkart to contain such sources and leveraged ChatGPT to annotate summaries using reviews and additional sources as input, amounting to $662$ opinion summaries in total. Statistics for the extended versions of the test sets are in Table \ref{Table: data}. For extensions, we obtain the additional sources (except for Flipkart) from the Amazon data \citep{amazondata}. We leverage ChatGPT as our annotator following recent works \citep{Gilardi_2023, Huang_2023}.  For each test set, we curated: \textbf{GPT-R}, in which summaries are generated using only reviews, and \textbf{GPT-RDQ}, in which summaries are generated using reviews, description, and question-answers.  We investigated multiple prompts before finalizing the best one (\textbf{Appendix \ref{prompt_appendix}}). We employed three professionals to evaluate the annotation quality on \textit{informativeness, faithfulness, coherence, conciseness,} and \textit{fluency} using a $5$-point scale. Statistics are in Table \ref{tab:annotation_quality}. The Inter-Rater Reliability computed using Fleiss' Kappa was $0.23, 0.41$ and $0.42$ for human-annotated, GPT-R, and GPT-RDQ summaries which are considered \textit{fair, moderate,} and \textit{moderate} agreement respectively \citep{landis1977measurement}. Refer to \textbf{Appendix \ref{eval_appendix}} \& \textbf{\ref{annotation_appendix}}. 


\subsection{Baseline Models}
\noindent\textbf{Extractive Approaches.} \textit{Random} selects a random review from the input as a lower bound. \textit{Oracle} is the extractive upper bound computed by selecting input sentences with the highest R1 to gold summary. \textit{Clustroid} \citep{brazinskas-etal-2020-unsupervised} selects the review with
the highest RL score with respect to other reviews. \textit{LexRank} \citep{Erkan2004LexRankGL} selects the most salient sentences from the input using BERT \citep{devlin-etal-2019-bert} encodings to represent sentences. \textit{QT} \citep{angelidis-etal-2021-extractive} represents opinions in quantized space.

\noindent\textbf{Abstractive Approaches.} \textit{CopyCat} \citep{brazinskas-etal-2020-unsupervised} is a hierarchical variational autoencoder that learns a latent code of the
summary. \textit{PlanSum} \citep{amplayo-lapata-2020-unsupervised} uses content plans to generate synthetic datasets. \textit{ConsistSum} \citep{Ke2022ConsistSumUO} uses aspect and sentiment distribution to generate review-summary pairs. \textit{MultimodalSum} \citep{im-etal-2021-self} generates summaries using multimodal data such as text, images, and meta-data. \textit{TransSum} \citep{wang-wan-2021-transsum} uses aspect and sentiment embeddings to construct synthetic datasets. \textit{COOP} \citep{iso-etal-2021-convex-aggregation} searches for convex combinations of latent vectors to generate summaries. \textit{AceSum} \citep{amplayo-etal-2021-aspect} uses silver-labeled data obtained through seed words to train the model. \textit{SW-LOO} \citep{shen-etal-2023-simple} uses the aspect seed words to construct synthetic datasets, whereas \textit{NLI-LOO} uses only aspects. \textit{Acesum\textsubscript{ext}}, \textit{SW-LOO\textsubscript{ext}}, and \textit{NLI-LOO\textsubscript{ext}} are the extractive versions respectively. \textit{ASBOS} \citep{Siledar2023AspectSentimentbasedOS} uses aspect-sentiment to filter sentences and generate supervised summaries.

\noindent\textbf{Multi-source Approaches.} Due to the absence of any unsupervised approaches that use additional sources as input we fine-tune two models using our synthetic dataset for a fair comparison. BART-concat and T$5$-concat use BART \citep{Lewis2019BARTDS} and T$5$ \citep{10.5555/3455716.3455856} respectively with the input as a concatenated text. \textbf{Appendix \ref{baseline_appendix}}.

\subsection{Implementation Details}
We used the \verb|bart-large| \citep{Lewis2019BARTDS} and \verb|t5-large| \citep{10.5555/3455716.3455856} models from HuggingFace \citep{Wolf2019TransformersSN}. 
A learning rate of $2e-6$, batch size of $8$, and $5$ epochs performs the best on dev sets (Appendix \ref{hyperparameter_appendix}). During inference, we set beam size to $5$ and no repeat ngram to $3$. For encoding, we use the \verb|all-MiniLM-L12-v2| from SBERT \citep{reimers-2019-sentence-bert}. For SDC, $k=8$ for Amazon and $10$ for Oposum+ and Flipkart, whereas top-p selection is done using $85$ percentile. Number of QA's are fixed to $M=10$. 

\begin{table}[t]
    \centering
    \resizebox{\columnwidth}{!}{%
    \begin{tabular}{p{\columnwidth}}
    \toprule
        \small{\textbf{Gold (Amazon GPT-RDQ)}}\\
    \midrule
    \small{The \textbf{VuPoint Film and Slide Scanner} presents a mixed picture. It claims ease \& efficiency for converting film to digital format, but user experiences differ. While some commend its user-friendliness and recommend it, others report frustrating issues like bleached images, compatibility problems, and subpar results. The need for XP compatibility limits its usefulness for modern systems. It offers potential for simple scanning but may require persistence to achieve desired outcomes.}\\
    \midrule
    \small{\textbf{MultimodalSum}}\\
    \midrule
    \small{I bought this product to scan my negatives. It does not work with Windows XP. I have tried to contact the company several times and have not received a response. I am very disappointed in the product. I would not recommend it to anyone.}\\
    \midrule
    \small{\textbf{BART-concat}}\\
    \midrule
    \small{I bought this Scanner to scan my slides and film. I have Windows XP and it does not work with it. \underline{I tried to download the drivers} from the web site but they are not available. I called the company and they said they would send me a new one but I have not received it yet. I am very disappointed.}\\
    \midrule
    \small{\textbf{Our Model (MEDOS)}}\\
    \midrule
    \small{I purchased the \textbf{VuPoint FS-C1-VP Film and Slide Digital Converter} to scan my \textbf{35mm film} and slide negatives. It is not compatible with Windows XP. \underline{The} \underline{software does not work with Windows 7 or 8}. I have tried to contact the company and they do not respond to my emails. I would not recommend this product to anyone.}\\
    \bottomrule
    \end{tabular}%
    }
    \caption{\textbf{Comparative Analysis.} ChatGPT-generated summary using \textit{reviews, description, and question-answers} (GPT-RDQ) followed by different model-generated summaries for an Amazon test set product. Information assisted from the description and question-answers are in \textbf{bold} and \underline{underline} respectively. MEDOS is able to capture vital information from additional sources which won't be possible using only reviews.} 
    \label{Table: example_summaries}
\end{table}

\section{Results and Analysis}\label{results}
\textbf{Automatic Evaluation.}\; We use the ROUGE-\{1,2,L\} F1 score  \cite{lin-2004-rouge} (R1, R2 \& RL) to assess the generated summary quality. Tables \ref{Table: amazon_results}, \ref{Table: oposum_results} \& \ref{Table: flipkart_results} present the results on Amazon and its variants, Oposum+ and its variants, and Flipkart and its variants respectively. In general, we observe that our MEDOS model performs better than baselines and outperforms MultimodalSum on all nine test sets. 
Better results on GPT-RDQ versions are expected as our model and these test sets use all sources for generating summaries. However, we observe that even on the original and GPT-R test sets our models perform much better. The reason for this we believe is that under the presence of multiple sources, our models are better at figuring out what information is essential and needs to be presented in the summary. Our approach to creating synthetic datasets plays a vital role in this. By showing the model the most relevant summary that takes into consideration all the sources, our models are able to learn better the task of opinion summarization as evidenced by the results. Next, almost for all cases, we observe that MEDOS performs better than the combination of simple concatenation approach and single encoder models (BART-concat \& T$5$-concat). The MEDOS model due to its multi-encoder framework is able to selectively choose relevant information from the product description and question-answers. Additionally, we observe that single encoder models encounter context limitations in most cases thereby being unable to leverage the additional sources fully.

\noindent\textbf{Qualitative Analysis.}\;
Table \ref{Table: dummy} presents the summary generated by our MEDOS model for four different products from the Amazon test set. Product description typically contains brand names as well as aspect-specifics. We observe that MEDOS excels at picking these specific names and including them in the generated summaries at appropriate places ensuring that the summaries are coherent. For example, \textbf{$\mathbf{35}$mm film} in product $1$ is an essential information that gets included in the summary. MEDOS also demonstrated the ability to pick relevant information from question-answers keeping the opinions being summarized in context. In product $4$, the MEDOS model additionally gathers the \underline{compatibility of Nixon Coolpix L$820$} and the \underline{weight of the product} from question-answers. Overall, MEDOS, due to its multi-encoder architecture and assistance from synthetic datasets during training learns to fuse relevant information well.


%

\noindent\textbf{Comparative Analysis.}\; Sample summaries generated by our model and some baselines on an Amazon test set product are shown in Table \ref{Table: example_summaries}. MultimodalSum uses reviews, images, and meta-data, whereas Gold (Amazon GPT-RDQ), BART-concat, and our models use reviews, product description, and question-answers. In comparison to MultimodalSum, which also uses product description as part of the meta-data, MEDOS is able to capture details better such as \textbf{VuPoint FS-C1-VP Film and Slide
Digital Converter} (brand name) and \textbf{$\mathbf{35}$mm film} (information present only in description). In the presence of QA, MEDOS is able to provide relevant additional context to the information present in reviews. It picks details about Windows $7$ and $8$ from question-answers to present it along with the Windows XP. Finally, MEDOS does a better job compared to BART-concat in capturing details which we intuit is due to its multi-encoder framework. Additionally, the overall retention of the consensus opinions from the reviews is unaffected.

\noindent \textbf{Error Analysis.}\; Unfortunately, our models are also prone to occasional hallucinations. For example, product $3$ in Table \ref{Table: dummy} mentions that an adaptor is needed to charge iPhone $4$s. Though, \textit{needing an adaptor for some models} is mentioned in question-answers and \textit{iPhone 4s} in reviews, there is no evidence of \textit{iPhone 4s needing an adaptor}. We attribute such hallucinations to treating brand names such as \textit{iPhone 4s}, \textit{iPhone 5,} etc. as same.

\noindent \textbf{Ablation Study.}\; Table \ref{Table: amazon_ablation_results} presents the ablation study of our MEDOS model in using different sources on the Amazon GPT-RDQ test set. 
Results indicate that the combination of all sources performs the best.
Intuitively, a higher score on Amazon GPT-RDQ summaries indicates that our model is leveraging the additional sources to generate more informative summaries. Without question-answers, we observe a $2$ R1 point drop whereas, without the description a $5$ R1 point drop. As expected, the utility of the description is higher than the question-answers. Descriptions contain aspect-specifics which help in enriching the summaries. In contrast, question-answers provide information related to specific queries about the product, which may or may not contribute to the overall summary.
The distinction is evident, as using only reviews and question-answers results in poorer performance compared to using only reviews and description.  
\begin{table}[t]
    \centering
    \resizebox{\columnwidth}{!}{
    \begin{tabular}{lccc}
    \toprule
    & \multicolumn{3}{c}{\textbf{Amazon GPT-RDQ}} \\
    \cmidrule(lr){2-4}
    & R1 $\uparrow$ & R2 $\uparrow$ & RL $\uparrow$ \\
    \midrule 
    \textbf{MEDOS} &&&\\
    \hspace{2em} w. Reviews + Description + QA & {$\mathbf{25.44}$} & {$\mathbf{4.16}$} & {$\mathbf{16.45}$} \\
    \hspace{2em} w. Reviews + Description  & \underline{$23.54$} & \underline{$2.43$} & \underline{$14.81$} \\
    \hspace{2em} w. Reviews + QA  & $20.05$ & $1.36$ & $12.90$   \\
    \hspace{2em} w. Reviews & $21.26$ & $2.22$ & $13.68$  \\
    \bottomrule
    \end{tabular}%
    }
    \caption{\textbf{Ablation study} on Amazon GPT-RDQ. The highest utility comes from adding the description. QA in the presence of reviews and description aids the best.}
    \label{Table: amazon_ablation_results}
\end{table}

\noindent \textbf{Human Evaluation.} Table \ref{Table: bws} shows the Best-Worst Scaling \citep{Louviere2015BestWorstST} results, assessing the quality of opinion summaries.
Six Masters' students aged $21$-$30$ evaluated the model-generated summaries on: \textit{faithfulness}, \textit{coherence}, \textit{conciseness}, and \textit{fluency}. Each evaluator assigned a score of +$1$ for best, -$1$ for worst, and $0$ for the remaining models. Final scores were computed by averaging the scores from all the evaluators. Notably, MEDOS achieved the best scores on all criteria.
\begin{table}[t]
    \centering
    \resizebox{\columnwidth}{!}{%
    \begin{tabular}{lcccc}
    \toprule
    \textbf{Amazon} & Faithfulness $\uparrow$ & Coherence $\uparrow$ & Conciseness $\uparrow$ & Fluency $\uparrow$ \\
    \midrule
    PlanSum & -$0.50$ & -$0.66$ & -$0.63$ & -$0.68$\\
    MultimodalSum & \underline{$0.17$} & \underline{$0.16$} & \underline{$0.22$} & \underline{$0.14$}\\
    BART-concat & $0.05$ & $0.08$ & $0.07$ & $0.10$\\
    \textbf{MEDOS} & {$\mathbf{0.21}$} & {$\mathbf{0.41}$}  & {$\mathbf{0.23}$} & {$\mathbf{0.50}$} \\
    \bottomrule
    \end{tabular}%
    }
    \caption{\textbf{Best-Worst Scaling.} MEDOS generated summaries received better scores on all four criteria in human evaluation using the best-worst scaling method.}
    \label{Table: bws}
\end{table}%

\noindent \textbf{SDC approach effectiveness.}
Our SDC approach selects the pseudo-summary based on description and QA first, followed by reviews. This ensures that the model sees relevant information during training thereby learning two things: 
picking of relevant information from additional sources and generating opinion summaries. Table \ref{Table: sdc_appendix} reports the results obtained using different SDC approaches. 

\begin{table}[t]
    \centering
    \resizebox{\columnwidth}{!}{%
    \begin{tabular}{lccc}
    \toprule
    & \multicolumn{3}{c}{\textbf{Amazon GPT-RDQ}} \\
    \cmidrule(lr){2-4}
    & R1 $\uparrow$ & R2 $\uparrow$ & RL $\uparrow$ \\
    \midrule
    Our approach & {$\mathbf{25.44}$} & {$\mathbf{4.16}$} & {$\mathbf{16.45}$} \\
    Using only reviews for selection & $21.36$ & $2.04$ & $13.86$ \\
    Random selection & $14.31$ & $0.48$ & $10.20$   \\
    \bottomrule
    \end{tabular}%
    }
    \caption{\textbf{SDC approach analysis.} Our approach that uses description and question-answers along with reviews for selecting pseudo-summary performs the best. 
    }
    \label{Table: sdc_appendix}
\end{table}

\begin{figure}[h]
    \centering
    \includegraphics[width=\columnwidth]{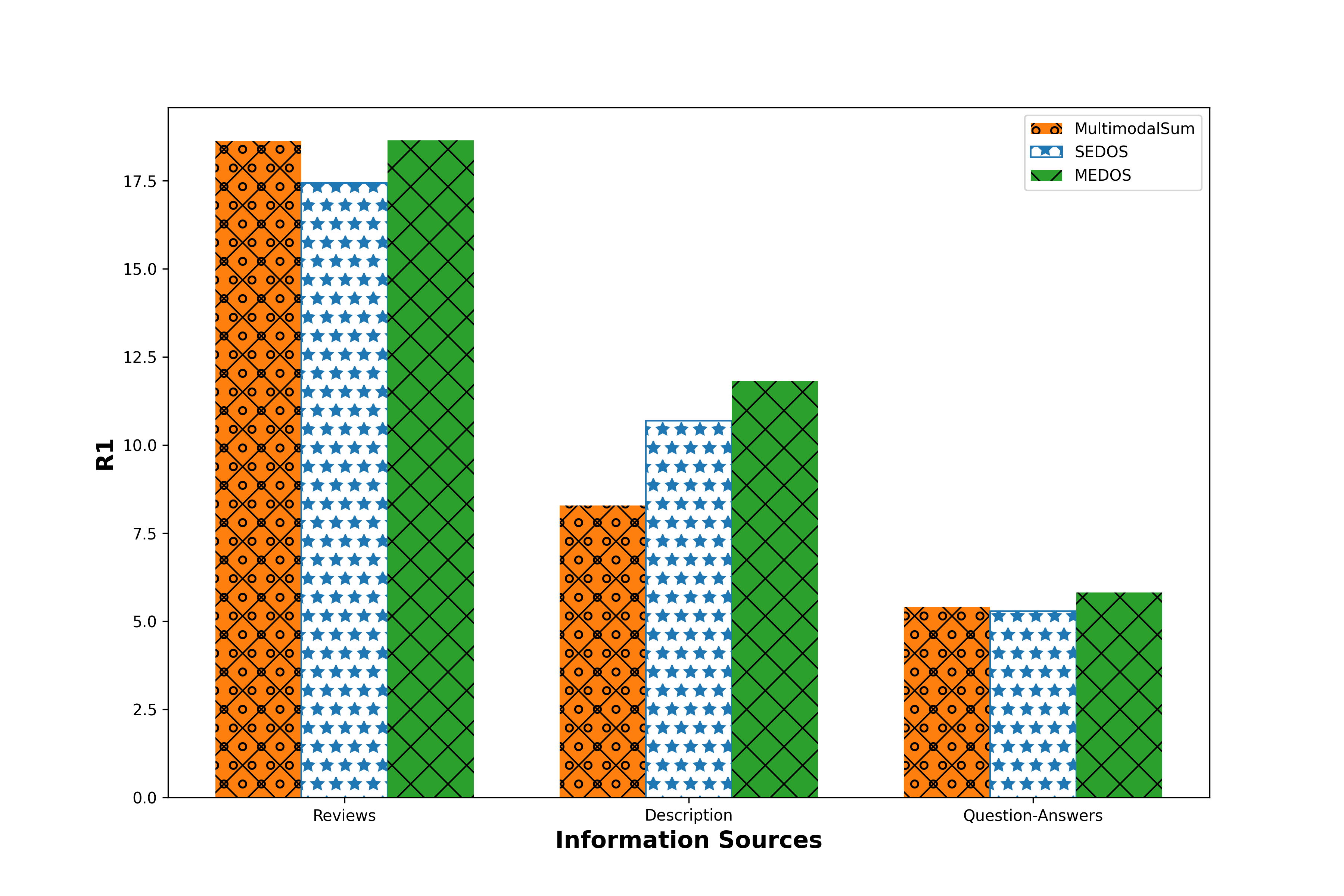}
    \caption{\textbf{Quantification of information captured.}
MEDOS captures a similar amount of information from
reviews as that of MultimodalSum, performs better for
description, and picks relevant details from QA.}
    \label{fig:quant_diag}
\end{figure}

\noindent \textbf{Quantification of information captured.}
We measure the R1 scores of generated summaries with the sources on the Amazon test set to quantify the amount of information captured. Figure \ref{fig:quant_diag} shows our MEDOS generated summaries achieve an R1 of $18.64$, $11.82$, and $5.81$ for reviews, description, and question-answers compared to $18.63$, $8.28$, and $5.46$ for MultimodalSum. The nearly identical R1 for MEDOS and MultimodalSum suggest that even when additional information is present, MEDOS effectively captures all the crucial details from reviews. 
Next, MEDOS is better than both MultimodalSum and BART-concat in leveraging the information from description. Finally, for QA, R1 for MultimodalSum acts as a baseline as it does not use any QA during summarization. We observe that the BART-concat performs worse whereas MEDOS is able to capture relevant information.

\noindent \textbf{MEDOS performance.}
We test the performance of MEDOS model
by varying the number of parameters. Specifically, we use two variants of BART i.e. \verb|bart-base| and \verb|bart-large|, and report the results in Table \ref{Table: medos_comp}. We observe that the \verb|bart-base| variant of the MEDOS with just $0.3$B parameters outperforms the single encoder models T$5$-concat and BART-concat (uses \verb|bart-large|). In comparison between the two variants of MEDOS, we find that the \verb|bart-large| version, as expected, performs better than \verb|bart-base| due to a larger number of parameters. Overall, our findings indicate that the multi-encoder performs better and is able to capture details from different sources effectively.

\begin{table}[t]
    \centering
    \resizebox{\columnwidth}{!}{%
    \begin{tabular}{lccccc}
    \toprule
    & & &\multicolumn{3}{c}{\textbf{Amazon GPT-RDQ}} \\
    \cmidrule(lr){4-6}
    & \textit{mul?} & \#parameters & R1 $\uparrow$ & R2 $\uparrow$ & RL $\uparrow$\\
    \midrule
    T$5$-concat & \xmark & $0.7$B & $20.61$ & $2.72$ & $13.33$\\
    BART-concat & \xmark & $0.4$B & $21.75$ & $2.39$ & $13.57$\\
    \midrule
    \textbf{MEDOS} &&&&&\\
    \hspace{2em} bart-base & \cmark & $0.3$B & \underline{$22.21$} & \underline{$3.38$} & \underline{$15.31$}\\
    \hspace{2em} bart-large & \cmark & $0.8$B & {$\mathbf{25.44}$} & {$\mathbf{4.16}$} & {$\mathbf{16.45}$}\\
    \bottomrule
    \end{tabular}%
    }
    \caption{\textbf{MEDOS Results.} Comparison of MEDOS summaries for different parameter sizes. \textit{mul?} represents models that use multiple encoders. \#parameters indicate the number of parameters in billions (B).}
    \label{Table: medos_comp}
\end{table}%

\noindent \textbf{LLMs on Multi-source Opinion Summarization.}
Recently, large language models (LLMs) have shown remarkable performance on a lot of tasks. For a fair comparison to baselines, we kept the focus of our work on smaller models in a self-supervised setting. For completion, we test the instruct models: Claude-$2$\footnote{\url{https://www.anthropic.com/index/claude-2}}, Chatglm$2$-$6$b \citep{du2022glm},  Llama-$2$-$70$b-chat\footnote{\url{https://huggingface.co/meta-llama/Llama-2-70b-chat-hf}}, and Llama-$2$-$7$b-chat\footnote{\url{https://huggingface.co/meta-llama/Llama-2-7b-chat-hf}} \citep{Touvron2023Llama2O} on the task of multi-source opinion summarization. The training details of these models are not public and could possibly had access to test sets as a part of their training. We use the same GPT-RDQ prompts as in Appendix \ref{prompt_appendix} to generate summaries using LLMs. We observe that our MEDOS model with just $0.8$B parameters performs comparably to Claude-$2$ with $130$B parameters and Chatglm-$6$b\footnote{\url{https://huggingface.co/THUDM/chatglm2-6b}} with $6$B parameters. Although Llama models with $70$B and $7$B parameters perform way better, for task-specific models MEDOS provides a cheaper alternative. 
\begin{table}[t]
    \centering
    \resizebox{\columnwidth}{!}{%
    \begin{tabular}{lcccc}
    \toprule
    & & \multicolumn{3}{c}{\textbf{Amazon GPT-RDQ}} \\
    \cmidrule(lr){3-5}
    \textbf{Model} & \#parameters & R1 $\uparrow$ & R2 $\uparrow$ & RL $\uparrow$ \\
    \midrule
    Claude-$2$ & $130$B & $31.11$ & $4.73$ & $16.67$ \\
    Llama-$2$-$70$b-chat & $70$B & {$\mathbf{32.77}$} & {$\mathbf{7.84}$} & {$\mathbf{20.28}$} \\
    Chatglm$2$-$6$b & $6$B & $27.31$ & $4.72$ & $16.80$ \\
    Llama-$2$-$7$b-chat & $7$B & \underline{$32.43$} & \underline{$7.33$} & \underline{$20.27$}\\
    \midrule
    \textbf{MEDOS} & $0.8$B & $25.44$ & $4.16$ & $16.45$ \\
    \bottomrule
    \end{tabular}%
    }
    \caption{\textbf{LLM results} on Amazon GPT-RDQ test set compared to MEDOS.} 
    \label{Table: llm_results}
\end{table}

\section{Conclusion and Future Work}
We proposed a novel approach to create synthetic datasets by harnessing information from \textbf{reviews} and additional sources such as \textbf{product description} and \textbf{question-answers}. This method enables supervised training of models without the necessity of expensive annotated training datasets.
Our proposed framework \textbf{MEDOS} uses separate encoders for selectively fusing information from these sources to generate an opinion summary.
For evaluation, due to the absence of any test sets that contained such additional sources and annotated summaries, we extended the already available e-commerce test sets with additional sources and leveraged ChatGPT to annotate summaries. This resulted in six additional test sets with $\mathbf{662}$ \textbf{opinion summaries} in total. Results show that our synthetic dataset approach and MEDOS framework outperforms the SOTA model on average by $\mathbf{14.5\%}$ and the simple input concatenation baseline by $\mathbf{6.5\%}$ across all nine test sets. Through qualitative and comparative analysis we demonstrated that our model-generated summaries are more informative and emphasize the importance of including additional sources for comprehensive summaries.

One future work is to expand these frameworks to encompass more reviews and all available sources, creating thorough product summaries.

\section*{Limitations}
Our work, although uses a multi-encoder framework, is still currently limited by the size of the input. In e-commerce, reviews generally tend to be in the tens of thousands which could not be supported directly by the current model architectures. There has been research on increasing the context limits of the latest large language models, however, the performance of such models needs to be tested in the context of handling larger inputs for the task of opinion summarization. It becomes even more challenging to integrate additional sources found on product pages on e-commerce websites to provide an overall well-rounded product summary. Finally, we did not consider large language models (LLMs) in our work as our goal was to push for improvements in smaller models for multi-source opinion summarization utilizing only the available product corpus without the need for expensive large-scale annotated datasets and compute-intensive large-scale models. Our models do not use any LLM signals or LLM-generated data for training and rely only on the product corpus for learning the task of multi-source opinion summarization.

\section*{Ethical Considerations}
We perform our experiments on existing opinion summarization datasets as well as extend the test sets by generating summaries using ChatGPT. Some of the examples in these datasets might not be appropriate for everyone. Our models may also propagate these unintended biases due to the nature of the datasets. We urge the research community to use our models and these test sets with caution and we are fully committed to removing any discrepancies in the existing datasets in the future.

\bibliography{custom,anthology}

\begin{thebibliography}{41}
\expandafter\ifx\csname natexlab\endcsname\relax\def\natexlab#1{#1}\fi

\bibitem[{Amplayo et~al.(2020)Amplayo, Angelidis, and Lapata}]{Amplayo2020UnsupervisedOS}
Reinald~Kim Amplayo, Stefanos Angelidis, and Mirella Lapata. 2020.
\newblock \href {https://api.semanticscholar.org/CorpusID:229156019} {Unsupervised opinion summarization with content planning}.
\newblock In \emph{AAAI Conference on Artificial Intelligence}.

\bibitem[{Amplayo et~al.(2021)Amplayo, Angelidis, and Lapata}]{amplayo-etal-2021-aspect}
Reinald~Kim Amplayo, Stefanos Angelidis, and Mirella Lapata. 2021.
\newblock \href {https://doi.org/10.18653/v1/2021.emnlp-main.528} {Aspect-controllable opinion summarization}.
\newblock In \emph{Proceedings of the 2021 Conference on Empirical Methods in Natural Language Processing}, pages 6578--6593, Online and Punta Cana, Dominican Republic. Association for Computational Linguistics.

\bibitem[{Amplayo and Lapata(2020)}]{amplayo-lapata-2020-unsupervised}
Reinald~Kim Amplayo and Mirella Lapata. 2020.
\newblock \href {https://doi.org/10.18653/v1/2020.acl-main.175} {Unsupervised opinion summarization with noising and denoising}.
\newblock In \emph{Proceedings of the 58th Annual Meeting of the Association for Computational Linguistics}, pages 1934--1945, Online. Association for Computational Linguistics.

\bibitem[{Angelidis et~al.(2021)Angelidis, Amplayo, Suhara, Wang, and Lapata}]{angelidis-etal-2021-extractive}
Stefanos Angelidis, Reinald~Kim Amplayo, Yoshihiko Suhara, Xiaolan Wang, and Mirella Lapata. 2021.
\newblock \href {https://doi.org/10.1162/tacl_a_00366} {Extractive opinion summarization in quantized transformer spaces}.
\newblock \emph{Transactions of the Association for Computational Linguistics}, 9:277--293.

\bibitem[{Angelidis and Lapata(2018)}]{angelidis-lapata-2018-summarizing}
Stefanos Angelidis and Mirella Lapata. 2018.
\newblock \href {https://doi.org/10.18653/v1/D18-1403} {Summarizing opinions: Aspect extraction meets sentiment prediction and they are both weakly supervised}.
\newblock In \emph{Proceedings of the 2018 Conference on Empirical Methods in Natural Language Processing}, pages 3675--3686, Brussels, Belgium. Association for Computational Linguistics.

\bibitem[{Bhaskar et~al.(2023)Bhaskar, Fabbri, and Durrett}]{bhaskar-etal-2023-prompted}
Adithya Bhaskar, Alex Fabbri, and Greg Durrett. 2023.
\newblock \href {https://doi.org/10.18653/v1/2023.findings-acl.591} {Prompted opinion summarization with {GPT}-3.5}.
\newblock In \emph{Findings of the Association for Computational Linguistics: ACL 2023}, pages 9282--9300, Toronto, Canada. Association for Computational Linguistics.

\bibitem[{Bra{\v{z}}inskas et~al.(2020)Bra{\v{z}}inskas, Lapata, and Titov}]{brazinskas-etal-2020-unsupervised}
Arthur Bra{\v{z}}inskas, Mirella Lapata, and Ivan Titov. 2020.
\newblock \href {https://doi.org/10.18653/v1/2020.acl-main.461} {Unsupervised opinion summarization as copycat-review generation}.
\newblock In \emph{Proceedings of the 58th Annual Meeting of the Association for Computational Linguistics}, pages 5151--5169, Online. Association for Computational Linguistics.

\bibitem[{Devlin et~al.(2019)Devlin, Chang, Lee, and Toutanova}]{devlin-etal-2019-bert}
Jacob Devlin, Ming-Wei Chang, Kenton Lee, and Kristina Toutanova. 2019.
\newblock \href {https://doi.org/10.18653/v1/N19-1423} {{BERT}: Pre-training of deep bidirectional transformers for language understanding}.
\newblock In \emph{Proceedings of the 2019 Conference of the North {A}merican Chapter of the Association for Computational Linguistics: Human Language Technologies, Volume 1 (Long and Short Papers)}, pages 4171--4186, Minneapolis, Minnesota. Association for Computational Linguistics.

\bibitem[{Du et~al.(2022)Du, Qian, Liu, Ding, Qiu, Yang, and Tang}]{du2022glm}
Zhengxiao Du, Yujie Qian, Xiao Liu, Ming Ding, Jiezhong Qiu, Zhilin Yang, and Jie Tang. 2022.
\newblock Glm: General language model pretraining with autoregressive blank infilling.
\newblock In \emph{Proceedings of the 60th Annual Meeting of the Association for Computational Linguistics (Volume 1: Long Papers)}, pages 320--335.

\bibitem[{Elsahar et~al.(2021)Elsahar, Coavoux, Rozen, and Gall{\'e}}]{elsahar-etal-2021-self}
Hady Elsahar, Maximin Coavoux, Jos Rozen, and Matthias Gall{\'e}. 2021.
\newblock \href {https://doi.org/10.18653/v1/2021.eacl-main.141} {Self-supervised and controlled multi-document opinion summarization}.
\newblock In \emph{Proceedings of the 16th Conference of the European Chapter of the Association for Computational Linguistics: Main Volume}, pages 1646--1662, Online. Association for Computational Linguistics.

\bibitem[{Erkan and Radev(2004)}]{Erkan2004LexRankGL}
G{\"u}nes Erkan and Dragomir~R. Radev. 2004.
\newblock Lexrank: Graph-based lexical centrality as salience in text summarization.
\newblock \emph{J. Artif. Intell. Res.}, 22:457--479.

\bibitem[{Gilardi et~al.(2023)Gilardi, Alizadeh, and Kubli}]{Gilardi_2023}
Fabrizio Gilardi, Meysam Alizadeh, and Maël Kubli. 2023.
\newblock \href {https://doi.org/10.1073/pnas.2305016120} {{ChatGPT} outperforms crowd workers for text-annotation tasks}.
\newblock \emph{Proceedings of the National Academy of Sciences}, 120(30).

\bibitem[{He and McAuley(2016)}]{amazondata}
Ruining He and Julian McAuley. 2016.
\newblock \href {https://doi.org/10.1145/2872427.2883037} {Ups and downs: Modeling the visual evolution of fashion trends with one-class collaborative filtering}.
\newblock In \emph{Proceedings of the 25th International Conference on World Wide Web}, WWW '16, page 507–517, Republic and Canton of Geneva, CHE. International World Wide Web Conferences Steering Committee.

\bibitem[{Hosking et~al.(2023)Hosking, Tang, and Lapata}]{hosking-etal-2023-attributable}
Tom Hosking, Hao Tang, and Mirella Lapata. 2023.
\newblock \href {https://doi.org/10.18653/v1/2023.acl-long.473} {Attributable and scalable opinion summarization}.
\newblock In \emph{Proceedings of the 61st Annual Meeting of the Association for Computational Linguistics (Volume 1: Long Papers)}, pages 8488--8505, Toronto, Canada. Association for Computational Linguistics.

\bibitem[{Hu and Liu(2006)}]{hu2006opinion}
Minqing Hu and Bing Liu. 2006.
\newblock Opinion extraction and summarization on the web.
\newblock In \emph{Aaai}, volume~7, pages 1621--1624.

\bibitem[{Huang et~al.(2023)Huang, Kwak, and An}]{Huang_2023}
Fan Huang, Haewoon Kwak, and Jisun An. 2023.
\newblock \href {https://doi.org/10.1145/3543873.3587368} {Is {ChatGPT} better topenai human annotators? potential and limitations of {ChatGPT} in explaining implicit hate speech}.
\newblock In \emph{Companion Proceedings of the {ACM} Web Conference 2023}. {ACM}.

\bibitem[{Im et~al.(2021)Im, Kim, Lee, Cho, and Chung}]{im-etal-2021-self}
Jinbae Im, Moonki Kim, Hoyeop Lee, Hyunsouk Cho, and Sehee Chung. 2021.
\newblock \href {https://doi.org/10.18653/v1/2021.acl-long.33} {Self-supervised multimodal opinion summarization}.
\newblock In \emph{Proceedings of the 59th Annual Meeting of the Association for Computational Linguistics and the 11th International Joint Conference on Natural Language Processing (Volume 1: Long Papers)}, pages 388--403, Online. Association for Computational Linguistics.

\bibitem[{Iso et~al.(2021)Iso, Wang, Suhara, Angelidis, and Tan}]{iso-etal-2021-convex-aggregation}
Hayate Iso, Xiaolan Wang, Yoshihiko Suhara, Stefanos Angelidis, and Wang-Chiew Tan. 2021.
\newblock \href {https://doi.org/10.18653/v1/2021.findings-emnlp.328} {{C}onvex {A}ggregation for {O}pinion {S}ummarization}.
\newblock In \emph{Findings of the Association for Computational Linguistics: EMNLP 2021}, pages 3885--3903, Punta Cana, Dominican Republic. Association for Computational Linguistics.

\bibitem[{Ke et~al.(2022)Ke, Gao, Shen, and Cheng}]{Ke2022ConsistSumUO}
Wenjun Ke, Jinhua Gao, Huawei Shen, and Xueqi Cheng. 2022.
\newblock Consistsum: Unsupervised opinion summarization with the consistency of aspect, sentiment and semantic.
\newblock \emph{Proceedings of the Fifteenth ACM International Conference on Web Search and Data Mining}.

\bibitem[{Kingma and Ba(2015)}]{Kingma2015AdamAM}
Diederik~P. Kingma and Jimmy Ba. 2015.
\newblock Adam: A method for stochastic optimization.
\newblock \emph{CoRR}, abs/1412.6980.

\bibitem[{Landis and Koch(1977)}]{landis1977measurement}
J~Richard Landis and Gary~G Koch. 1977.
\newblock The measurement of observer agreement for categorical data.
\newblock \emph{biometrics}, pages 159--174.

\bibitem[{Lewis et~al.(2019)Lewis, Liu, Goyal, Ghazvininejad, rahman Mohamed, Levy, Stoyanov, and Zettlemoyer}]{Lewis2019BARTDS}
Mike Lewis, Yinhan Liu, Naman Goyal, Marjan Ghazvininejad, Abdel rahman Mohamed, Omer Levy, Veselin Stoyanov, and Luke Zettlemoyer. 2019.
\newblock Bart: Denoising sequence-to-sequence pre-training for natural language generation, translation, and comprehension.
\newblock In \emph{Annual Meeting of the Association for Computational Linguistics}.

\bibitem[{Li et~al.(2020)Li, Yuan, Xu, Wu, He, and Zhou}]{Li_Yuan_Xu_Wu_He_Zhou_2020}
Haoran Li, Peng Yuan, Song Xu, Youzheng Wu, Xiaodong He, and Bowen Zhou. 2020.
\newblock \href {https://doi.org/10.1609/aaai.v34i05.6332} {Aspect-aware multimodal summarization for chinese e-commerce products}.
\newblock \emph{Proceedings of the AAAI Conference on Artificial Intelligence}, 34(05):8188--8195.

\bibitem[{Lin(2004)}]{lin-2004-rouge}
Chin-Yew Lin. 2004.
\newblock \href {https://aclanthology.org/W04-1013} {{ROUGE}: A package for automatic evaluation of summaries}.
\newblock In \emph{Text Summarization Branches Out}, pages 74--81, Barcelona, Spain. Association for Computational Linguistics.

\bibitem[{Liu and Lapata(2019)}]{Liu2019TextSW}
Yang Liu and Mirella Lapata. 2019.
\newblock Text summarization with pretrained encoders.
\newblock \emph{ArXiv}, abs/1908.08345.

\bibitem[{Louviere et~al.(2015)Louviere, Flynn, and Marley}]{Louviere2015BestWorstST}
Jordan~J. Louviere, Terry~N. Flynn, and Anthony A.~J. Marley. 2015.
\newblock Best-worst scaling: Theory, methods and applications.

\bibitem[{Nallapati et~al.(2016)Nallapati, Zhou, dos Santos, Çaglar G{\"u}lçehre, and Xiang}]{Nallapati2016AbstractiveTS}
Ramesh Nallapati, Bowen Zhou, C{\'i}cero~Nogueira dos Santos, Çaglar G{\"u}lçehre, and Bing Xiang. 2016.
\newblock Abstractive text summarization using sequence-to-sequence rnns and beyond.
\newblock In \emph{Conference on Computational Natural Language Learning}.

\bibitem[{{OpenAI}(2023)}]{openai2023}
{OpenAI}. 2023.
\newblock {ChatGPT (August 3 Version)}.
\newblock \url{https://chat.openai.com}.

\bibitem[{Raffel et~al.(2020)Raffel, Shazeer, Roberts, Lee, Narang, Matena, Zhou, Li, and Liu}]{10.5555/3455716.3455856}
Colin Raffel, Noam Shazeer, Adam Roberts, Katherine Lee, Sharan Narang, Michael Matena, Yanqi Zhou, Wei Li, and Peter~J. Liu. 2020.
\newblock Exploring the limits of transfer learning with a unified text-to-text transformer.
\newblock \emph{J. Mach. Learn. Res.}, 21(1).

\bibitem[{Reimers and Gurevych(2019)}]{reimers-2019-sentence-bert}
Nils Reimers and Iryna Gurevych. 2019.
\newblock \href {https://arxiv.org/abs/1908.10084} {Sentence-bert: Sentence embeddings using siamese bert-networks}.
\newblock In \emph{Proceedings of the 2019 Conference on Empirical Methods in Natural Language Processing}. Association for Computational Linguistics.

\bibitem[{Rush et~al.(2015)Rush, Chopra, and Weston}]{rush-etal-2015-neural}
Alexander~M. Rush, Sumit Chopra, and Jason Weston. 2015.
\newblock \href {https://doi.org/10.18653/v1/D15-1044} {A neural attention model for abstractive sentence summarization}.
\newblock In \emph{Proceedings of the 2015 Conference on Empirical Methods in Natural Language Processing}, pages 379--389, Lisbon, Portugal. Association for Computational Linguistics.

\bibitem[{See et~al.(2017)See, Liu, and Manning}]{See2017GetTT}
A.~See, Peter~J. Liu, and Christopher~D. Manning. 2017.
\newblock Get to the point: Summarization with pointer-generator networks.
\newblock \emph{ArXiv}, abs/1704.04368.

\bibitem[{Shen et~al.(2023)Shen, Ma, Wang, Vyas, Dixit, Ballesteros, and Benajiba}]{shen-etal-2023-simple}
Ming Shen, Jie Ma, Shuai Wang, Yogarshi Vyas, Kalpit Dixit, Miguel Ballesteros, and Yassine Benajiba. 2023.
\newblock \href {https://aclanthology.org/2023.findings-eacl.142} {Simple yet effective synthetic dataset construction for unsupervised opinion summarization}.
\newblock In \emph{Findings of the Association for Computational Linguistics: EACL 2023}, pages 1898--1911, Dubrovnik, Croatia. Association for Computational Linguistics.

\bibitem[{Siledar et~al.(2023{\natexlab{a}})Siledar, Banerjee, Patil, Singh, Chelliah, Garera, and Bhattacharyya}]{siledar-etal-2023-synthesize}
Tejpalsingh Siledar, Suman Banerjee, Amey Patil, Sudhanshu Singh, Muthusamy Chelliah, Nikesh Garera, and Pushpak Bhattacharyya. 2023{\natexlab{a}}.
\newblock \href {https://doi.org/10.18653/v1/2023.findings-emnlp.899} {Synthesize, if you do not have: Effective synthetic dataset creation strategies for self-supervised opinion summarization in {E}-commerce}.
\newblock In \emph{Findings of the Association for Computational Linguistics: EMNLP 2023}, pages 13480--13491, Singapore. Association for Computational Linguistics.

\bibitem[{Siledar et~al.(2023{\natexlab{b}})Siledar, Makwana, and Bhattacharyya}]{Siledar2023AspectSentimentbasedOS}
Tejpalsingh Siledar, Jigar Makwana, and Pushpak Bhattacharyya. 2023{\natexlab{b}}.
\newblock Aspect-sentiment-based opinion summarization using multiple information sources.
\newblock \emph{Proceedings of the 6th Joint International Conference on Data Science \& Management of Data (10th ACM IKDD CODS and 28th COMAD)}.

\bibitem[{Touvron et~al.(2023)Touvron, Martin, Stone, Albert, Almahairi, Babaei, Bashlykov, Batra, Bhargava, Bhosale, Bikel, Blecher, Ferrer, Chen, Cucurull, Esiobu, Fernandes, Fu, Fu, Fuller, Gao, Goswami, Goyal, Hartshorn, Hosseini, Hou, Inan, Kardas, Kerkez, Khabsa, Kloumann, Korenev, Koura, Lachaux, Lavril, Lee, Liskovich, Lu, Mao, Martinet, Mihaylov, Mishra, Molybog, Nie, Poulton, Reizenstein, Rungta, Saladi, Schelten, Silva, Smith, Subramanian, Tan, Tang, Taylor, Williams, Kuan, Xu, Yan, Zarov, Zhang, Fan, Kambadur, Narang, Rodriguez, Stojnic, Edunov, and Scialom}]{Touvron2023Llama2O}
Hugo Touvron, Louis Martin, Kevin~R. Stone, Peter Albert, Amjad Almahairi, Yasmine Babaei, Nikolay Bashlykov, Soumya Batra, Prajjwal Bhargava, Shruti Bhosale, Daniel~M. Bikel, Lukas Blecher, Cristian~Cant{\'o}n Ferrer, Moya Chen, Guillem Cucurull, David Esiobu, Jude Fernandes, Jeremy Fu, Wenyin Fu, Brian Fuller, Cynthia Gao, Vedanuj Goswami, Naman Goyal, Anthony~S. Hartshorn, Saghar Hosseini, Rui Hou, Hakan Inan, Marcin Kardas, Viktor Kerkez, Madian Khabsa, Isabel~M. Kloumann, A.~V. Korenev, Punit~Singh Koura, Marie-Anne Lachaux, Thibaut Lavril, Jenya Lee, Diana Liskovich, Yinghai Lu, Yuning Mao, Xavier Martinet, Todor Mihaylov, Pushkar Mishra, Igor Molybog, Yixin Nie, Andrew Poulton, Jeremy Reizenstein, Rashi Rungta, Kalyan Saladi, Alan Schelten, Ruan Silva, Eric~Michael Smith, R.~Subramanian, Xia Tan, Binh Tang, Ross Taylor, Adina Williams, Jian~Xiang Kuan, Puxin Xu, Zhengxu Yan, Iliyan Zarov, Yuchen Zhang, Angela Fan, Melanie Kambadur, Sharan Narang, Aurelien Rodriguez, Robert Stojnic, Sergey Edunov, and
  Thomas Scialom. 2023.
\newblock \href {https://api.semanticscholar.org/CorpusID:259950998} {Llama 2: Open foundation and fine-tuned chat models}.
\newblock \emph{ArXiv}, abs/2307.09288.

\bibitem[{Vaswani et~al.(2017)Vaswani, Shazeer, Parmar, Uszkoreit, Jones, Gomez, Kaiser, and Polosukhin}]{NIPS2017_3f5ee243}
Ashish Vaswani, Noam Shazeer, Niki Parmar, Jakob Uszkoreit, Llion Jones, Aidan~N Gomez, \L~ukasz Kaiser, and Illia Polosukhin. 2017.
\newblock \href {https://proceedings.neurips.cc/paper_files/paper/2017/file/3f5ee243547dee91fbd053c1c4a845aa-Paper.pdf} {Attention is all you need}.
\newblock In \emph{Advances in Neural Information Processing Systems}, volume~30. Curran Associates, Inc.

\bibitem[{Wang and Wan(2021)}]{wang-wan-2021-transsum}
Ke~Wang and Xiaojun Wan. 2021.
\newblock \href {https://doi.org/10.18653/v1/2021.findings-acl.65} {{T}rans{S}um: Translating aspect and sentiment embeddings for self-supervised opinion summarization}.
\newblock In \emph{Findings of the Association for Computational Linguistics: ACL-IJCNLP 2021}, pages 729--742, Online. Association for Computational Linguistics.

\bibitem[{Wang and Ling(2016)}]{wang-ling-2016-neural}
Lu~Wang and Wang Ling. 2016.
\newblock \href {https://doi.org/10.18653/v1/N16-1007} {Neural network-based abstract generation for opinions and arguments}.
\newblock In \emph{Proceedings of the 2016 Conference of the North {A}merican Chapter of the Association for Computational Linguistics: Human Language Technologies}, pages 47--57, San Diego, California. Association for Computational Linguistics.

\bibitem[{Wolf et~al.(2019)Wolf, Debut, Sanh, Chaumond, Delangue, Moi, Cistac, Rault, Louf, Funtowicz, and Brew}]{Wolf2019TransformersSN}
Thomas Wolf, Lysandre Debut, Victor Sanh, Julien Chaumond, Clement Delangue, Anthony Moi, Pierric Cistac, Tim Rault, R{\'e}mi Louf, Morgan Funtowicz, and Jamie Brew. 2019.
\newblock Transformers: State-of-the-art natural language processing.
\newblock In \emph{Conference on Empirical Methods in Natural Language Processing}.

\bibitem[{Zhao and Chaturvedi(2020)}]{zhao2020weakly}
Chao Zhao and Snigdha Chaturvedi. 2020.
\newblock Weakly-supervised opinion summarization by leveraging external information.
\newblock In \emph{Proceedings of the AAAI Conference on Artificial Intelligence}, volume~34, pages 9644--9651.

\end{thebibliography}

\appendix

\section{Results on Oposum+ and Flipkart datasets}
Results on Oposum+ and Flipkart and their corresponding extended test sets are reported in Tables \ref{Table: oposum_results} and \ref{Table: flipkart_results} respectively.
\begin{table*}[h]
    \centering
    \small
    \resizebox{2\columnwidth}{!}{%
    \begin{tabular}{clcccccccccccc}
    \toprule
    &&&&& \multicolumn{3}{c}{\textbf{Oposum+}} & \multicolumn{3}{c}{\textbf{Oposum+ GPT-R}} & \multicolumn{3}{c}{\textbf{Oposum+ GPT-RDQ}} \\
    \cmidrule(lr){6-8}\cmidrule(lr){9-11}\cmidrule(lr){12-14}
    \textit{abs?}&\textbf{Model} &\textbf{R}& \textbf{D}& \textbf{Q} & R1 $\uparrow$ & R2 $\uparrow$ & RL $\uparrow$ & R1 $\uparrow$ & R2 $\uparrow$ & RL $\uparrow$ & R1 $\uparrow$ & R2 $\uparrow$ & RL $\uparrow$ \\
    \midrule
     \xmark &Random  &\cmark & \xmark & \xmark  & $33.63$ & $10.79$ & $19.82$   & $24.08$ & $2.38$ & $13.25$  & $23.68$ & $2.12$ & $12.98$   \\
     \xmark &Oracle  &\cmark & \xmark & \xmark  & $77.31$ & $70.30$  & $74.35$  & $36.87$ & $7.41$ & $23.88$  & $36.28$ & $7.44$ & $23.87$   \\
     \midrule
    \xmark
    &QT &\cmark & \cmark & \cmark & $37.72$ & $14.65$ & $21.69$    & $25.82$ & $3.47$ & $14.01$ & $25.81$ & $3.21$ & $14.13$    \\
    \xmark &AceSum\textsubscript{ext} &\cmark & \xmark & \xmark & $38.48$ & $15.17$ & $22.82$ &  - & - & -   & - & - & -   \\
    \xmark &SW-LOO\textsubscript{ext} &\cmark & \xmark & \xmark & $40.45$ & $19.13$  & $23.20$  & - & - & - & - & - & -  \\
    \xmark &NLI-LOO\textsubscript{ext} &\cmark & \xmark & \xmark & $39.79$ & $18.33$  & $23.49$  & - & - & - & - & - & -  \\
    \midrule
    \cmark &CopyCat  &\cmark & \xmark & \xmark       & $29.80$ & $5.61$ & $17.97$ & $22.41$ & $2.30$ & $13.94$ & $22.38$ & $2.03$ & $14.06$    \\
    \cmark &AceSum &\cmark & \xmark & \xmark & $32.98$ & $10.72$ & $20.27$ &  $22.78$ & $3.59$ & $13.20$   & $23.54$ & $3.51$ & $13.88$   \\
    \cmark &PlanSum  &\cmark & \xmark & \xmark  & $30.26$  & $5.29$  & $17.48$   & $22.37$ & $2.05$ &  $13.32$ & $22.64$ & $2.25$ & $13.71$     \\
    \cmark &MultimodalSum &\cmark & \cmark & \xmark & $33.08$ & $7.46$  & $19.75$  & $23.35$ & $2.98$ & $14.53$ & $23.73$ & $2.80$ & $14.70$  \\
    \cmark &SW-LOO &\cmark & \xmark & \xmark & \underline{$36.19$} & {$\mathbf{12.17}$}  & \underline{$21.11$}  & - & - & - & - & - & -  \\
    \cmark &NLI-LOO &\cmark & \xmark & \xmark & $31.22$ & $9.93$  & $19.08$  & - & - & - & - & - & -  \\

    \midrule
    \cmark& T$5$-concat &\cmark & \cmark & \cmark  & $30.84$  & \underline{$11.08$}  & 
    $21.01$   &   $21.98$ & $2.84$ & $12.91$ & $20.41$ & $2.31$ & $12.73$ \\
    \cmark& BART-concat &\cmark & \cmark & \cmark  & $34.76$  & $9.12$  & 
    $20.64$   &   \underline{$25.64$} & \underline{$3.47$} & \underline{$15.29$} & \underline{$25.62$} & {$\mathbf{3.36}$} & \underline{$15.91$} \\

    \midrule
    \cmark& \textbf{MEDOS} &\cmark & \cmark & \cmark  & {$\mathbf{36.57}$*}  & $8.79$*  & {$\mathbf{21.35}$*}   &   {$\mathbf{26.82}$*} & {$\mathbf{3.67}$*} & {$\mathbf{15.92}$*} & {$\mathbf{26.32}$*} & \underline{$3.34$}* & {$\mathbf{16.10}$*} \\
    \bottomrule
    \end{tabular}%
    }
    \caption{\textbf{Results on Oposum+ test set and its extensions.} \textbf{R}, \textbf{D}, \textbf{Q} indicate the presence of \textit{reviews}, \textit{description}, and \textit{question-answers} respectively in the input. \textit{abs?} indicate abstractive systems. \textbf{Bold} and \underline{underline} indicate best and second-best scores using abstractive systems. * indicates pvalue $<0.05$ on paired t-test against MultimodalSum. Overall our combination of SDC approach and MEDOS model outperforms baselines across all three test sets.}
    \label{Table: oposum_results}
\end{table*}%
\begin{table*}[h]
    \centering
    \small
    \resizebox{2\columnwidth}{!}{%
    \begin{tabular}{clcccccccccccc}
    \toprule
    &&&&& \multicolumn{3}{c}{\textbf{Flipkart}} & \multicolumn{3}{c}{\textbf{Flipkart GPT-R}} & \multicolumn{3}{c}{\textbf{Flipkart GPT-RDQ}} \\
    \cmidrule(lr){6-8}\cmidrule(lr){9-11}\cmidrule(lr){12-14}
    \textit{abs?} & \textbf{Model} &\textbf{R}& \textbf{D}& \textbf{Q} & R1 $\uparrow$ & R2 $\uparrow$ & RL $\uparrow$  & R1 $\uparrow$ & R2 $\uparrow$ & RL $\uparrow$ & R1 $\uparrow$ & R2 $\uparrow$ & RL $\uparrow$\\
    \midrule
     \xmark &Random  &\cmark & \xmark & \xmark  & $19.50$ & $2.50$ & $10.89$   & $24.22$ & $4.40$ & $14.10$  & $18.04$ & $2.26$ & $10.51$   \\
     \xmark &Oracle  &\cmark & \xmark & \xmark  & $34.07$ & $6.34$  & $21.30$  &  $38.35$ & $9.98$  & $24.81$  & $29.47$ & $5.12$ & $19.20$   \\
    \midrule
     \xmark &Clustroid  &\cmark & \cmark & \cmark  & $21.42$ & $3.01$ & $12.08$   & $27.76$ & $5.56$ & $16.77$  & $10.17$ & $1.45$ & $7.74$   \\
    \xmark &LexRank  &\cmark & \cmark & \cmark        & $21.57$ & $2.66$ & $11.88$ & $28.19$ & $5.91$ & $16.92$ & $19.65$ & $3.03$ & $12.15$   \\
    \xmark &QT  &\cmark & \cmark & \cmark & $25.18$ & $3.62$ & $13.05$    & $30.94$ & $5.96$ & $15.34$ & $22.92$ & $2.95$ & $11.97$    \\
    \midrule
        \cmark &ASBOS$^\dagger$ &\cmark & \cmark & \cmark  & $32.55$ & $6.44$  & $17.03$ & $28.27$ & $4.05$ & $14.30$ & $27.32$ & $4.95$ & $14.83$  \\
    \midrule
    \cmark &CopyCat   &\cmark & \xmark & \xmark       & $18.38$ & $1.81$ & $11.99$ & $21.68$ & $2.13$ & $13.92$ & $17.84$ & $1.25$ & $11.70$    \\
    \cmark &PlanSum &\cmark & \xmark & \xmark    &   $19.96$ & $2.70$  & $12.86$   & $21.17$ & $2.23$ & $13.48$  & $17.34$ & $1.49$ & $11.68$      \\
    \cmark &MultimodalSum &\cmark & \cmark & \xmark  & $21.76$ & $3.23$  & $13.57$  & $23.60$ & $2.78$ & $15.01$ & $19.04$ & $1.79$ & $12.24$  \\
    \midrule
    \cmark & T$5$-concat &\cmark & \cmark & \cmark    & $20.41$  & $2.83$  & 
    $11.80$   &   \underline{$26.70$} & {$\mathbf{5.75}$} & \underline{$16.65$} & $20.14$ & $3.00$ & $12.31$ \\
    \cmark & BART-concat &\cmark & \cmark & \cmark    & \underline{$22.35$}  & \underline{$4.46$}  & 
    \underline{$15.53$}   &   {$\mathbf{27.27}$} & \underline{$4.51$} & {$\mathbf{17.22}$} & \underline{$23.29$} & \underline{$3.13$} & \underline{$14.98$} \\
    \midrule
    \cmark & \textbf{MEDOS} &\cmark & \cmark & \cmark   & {$\mathbf{25.97}$}*  & {$\mathbf{5.29}$}*  & {$\mathbf{16.05}$*}   &   $26.29$* & $4.03$* & $16.59$* & {$\mathbf{23.92}$*} & {$\mathbf{4.30}$*} & {$\mathbf{16.35}$*} \\
    \bottomrule
    \end{tabular}%
    }
    \caption{\textbf{Results on Flipkart test set and its extensions.} \textbf{R}, \textbf{D}, \textbf{Q} indicate the presence of \textit{reviews}, \textit{description}, and \textit{question-answers} respectively in the input. \textit{abs?} indicate abstractive systems. \textbf{Bold} and \underline{underline} indicate best and second-best using abstractive systems. * indicates pvalue $<0.05$ on paired t-test against MultimodalSum. $\dagger$ represents supervised systems. Overall our combination of SDC approach and MEDOS outperforms baselines.}
    \label{Table: flipkart_results}
\end{table*}%



%

\begin{table*}
    \centering
    \small
    \resizebox{2\columnwidth}{!}{%
    \begin{tabular}{lccccc}
    \toprule
     \textbf{Rating} & \textbf{1} & \textbf{2} & \textbf{3} & \textbf{4} & \textbf{5}\\
     \midrule
    Informativeness & very poor & poor & acceptable & good & very good\\
    Faithfulness & all hallucinated & somewhat verifiable & moderate hallucination & slight hallucination & no hallucination\\
    Coherence & very poor & poor & acceptable & good & very good\\
    Conciseness & verbose & moderately verbose & slightly verbose & almost concise & concise\\
    Fluency & ungrammatical & slightly fluent & somewhat fluent & mostly fluent & fluent\\
    \bottomrule
    \end{tabular}
    }
\caption{\textbf{Human evaluation metrics.} We use a scale of 1-5 to rate summaries on five evaluation metrics.}
\label{tab:likert}
\end{table*}

\section{GPT Prompts}\label{prompt_appendix}
\hangindent=2em
\hangafter=1
\noindent \textbf{GPT-R prompt}: 
\textit{Following are the reviews for a product. Generate a summary of the opinions as a review itself with a word limit of under 100 words. Use information from the given reviews only to generate the summary. \\
\textbf{reviews}: [r\textsubscript{1},...,r\textsubscript{k}]
}

\hangindent=2em
\hangafter=1
\noindent\textbf{GPT-RDQ prompt}:
\textit{Following are the reviews, description, and question-answers for a product. Generate a summary of the opinions as a review itself with a word limit of under 100 words. Use information from the given reviews, description, and question-answers only to generate the summary. \\
\textbf{reviews}: [r\textsubscript{1},...,r\textsubscript{k}]\\
\textbf{description} : "..."\\
\textbf{question-answers}: [q\textsubscript{1},..,q\textsubscript{M}]
}

\section{Evaluation Metric}\label{eval_appendix}
We use various metrics to qualitatively evaluate our model-generated summaries as well as ChatGPT-annotated summaries. We use the following:
\begin{enumerate}
    \item \textbf{Informativeness}- how much of the information is captured?
    \item \textbf{Faithfulness}- how consistent are the opinions compared to reference summaries?
    \item \textbf{Coherence}- is the summary well organized and easy to read? 
    \item \textbf{Conciseness}- is the summary concise yet informative?
    \item \textbf{Fluency}- is the summary fluent and grammatical?
\end{enumerate}

\section{ChatGPT Annotation Quality}
We assessed the GPT-generated summaries against human-written summaries on $5$ metrics namely Informativeness, Faithfulness, Coherence, Conciseness, and Fluency. Results are presented in Table \ref{tab:annotation_quality}. We compare the ChatGPT-generated summaries against the human-annotated summaries for different test sets and report the results in Table \ref{Table: rouge_overlap}. For ChatGPT-generated summaries refer to Table \ref{Table: example_summaries_appendix}. GPT-R represents ChatGPT summaries using only reviews as input whereas GPT-RDQ represents ChatGPT summaries using reviews, description and question-answers.

\begin{table}[h]
    \centering
    \resizebox{\columnwidth}{!}{%
    \begin{tabular}{lccccc}
    \toprule
    & & \multicolumn{3}{c}{\textbf{ChatGPT generated}} \\
    \cmidrule(lr){3-5}
    & No. of summaries & R1 $\uparrow$ & R2 $\uparrow$ & RL $\uparrow$\\
    \midrule
    Amazon & $96$ & $25.09$ & $2.58$ & $14.02$\\
    Oposum+ & $90$ & $30.01$ & $4.42$ & $15.30$\\
    Flipkart & $145$ & $30.20$ & $4.18$ & $15.74$\\
    \bottomrule
    \end{tabular}%
    }
    \caption{\textbf{ChatGPT Results.} Comparison of ChatGPT summaries with human-annotated summaries for different test sets.}
    \label{Table: rouge_overlap}
\end{table}%

\begin{table}[h]
    \centering
    \small
    \resizebox{\columnwidth}{!}{%
    \begin{tabular}{lccccc}
    \toprule
     &Info. $\uparrow$ & Faith. $\uparrow$ & Coh. $\uparrow$ & Con. $\uparrow$ & Flu. $\uparrow$\\
     \midrule
    Human & $3.88$ & $3.91$ & $3.68$ & $3.83$ & $3.62$\\
    GPT-R & $4.02$ & $4.13$ & $4.02$ & $4.09$ & $3.98$\\
    GPT-RDQ & $4.10$ & $4.16$ & $4.16$ & $4.23$ & $4.16$ \\
    \bottomrule
    \end{tabular}
    }
\caption{\textbf{Annotation quality.} Both GPT-R and GPT-RDQ summaries score higher on all the metrics on average compared to human-annotated summaries. Scores range from 1-5. Info-\textit{informativeness}, Faith-\textit{faithfulness}, Coh-\textit{coherence}, Con-\textit{conciseness}, Flu-\textit{fluency}.}
\label{tab:annotation_quality}
\end{table}

\section{Dataset Details}\label{dataset_appendix}
\hangindent=2em
\hangafter=1
\noindent\textbf{Amazon} Amazon contains reviews from $4$ domains: \textit{electronics, home \& kitchen, personal care,} and \textit{clothing, shoes \& jewelry}. 
The evaluation set contains $3$ summaries and $8$ reviews per product. The training set contains $\sim1$M reviews over $90$K products.

\hangindent=2em
\hangafter=1
\noindent\textbf{Oposum+} Oposum+ contains reviews from $6$ domains: \textit{bags, bluetooth headsets, boots, keyboards, televisions}. The evaluation set contains $3$ extractive summaries and $10$ reviews per product. The training set contains $\sim4.13$M reviews over $95$K products.

\hangindent=2em
\hangafter=1
\noindent
\textbf{Flipkart}\; Flipkart contains reviews from $3$ domains: \textit{laptops, mobiles}, and \textit{tablets}. The test set has $1$ summary per product. The original test set contains $1$K reviews per product on average. We downsample this to $10$ reviews per product (randomly) for comparison.

\section{Single-Encoder  Baseline}\label{baseline_appendix}
\begin{figure}[]
    \centering
    \includegraphics[width=\columnwidth]{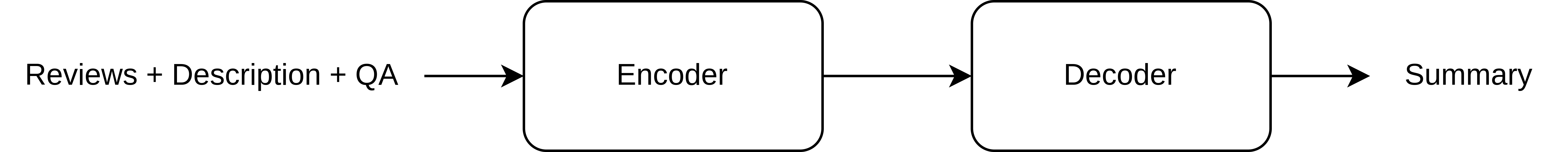}
    \caption{Framework of the baseline model that takes \textit{reviews}, \textit{description}, and \textit{QA} as the input. A simple concatenation (+) of the input sources is used to generate a summary. During inference, the model generates a summary whereas during training the model uses pseudo-summary obtained through SDC process for learning.} 
    \label{fig:single_enc_appendix}
\end{figure}
In the single-encoder framework, we concatenate reviews, product description, and question-answers using a separator symbol (</s>). This concatenated text $c_{rdq}$ goes through an encoder to get the fused attention $\mathbf{a_f}$ as:
\begin{align}
    \mathbf{a_f} &= \text{S-Attn}(\text{Enc}(c_{rdq}))
\end{align}
During training, the summary will be the pseudo-summary $r$ and the input $c_{rdq}$ will be formed using $T, D, Q$ from the synthetic quadruplet. Figure \ref{fig:single_enc_appendix} describes the single-encoder architecture. We use BART and T$5$ as our baseline models. 

\section{Implementation Details}\label{hyperparameter_appendix}
We used the Adam \citep{Kingma2015AdamAM} optimizer with eps of $1e-4$ and linear weight decay to optimize our models. We use learning rate in [$1e-6, 2e-6, 1e-5, 2e-5$] and batch size in [$8,16$] as our hyperparameters. All experiments use NVIDIA A100-SXM4-80GB GPUs.

\section{Inter-Rater Reliability}\label{annotation_appendix}
We employed three professionals proficient in English in the age group of 23-34. Two evaluators were male and one was female. They were provided with detailed evaluation instructions along with examples to rate summaries on different criteria as shown in Table \ref{tab:likert}. Each instance of the dataset was rated once and the work was equally divided among the three evaluators. 100 summaries were randomly chosen for evaluation and each evaluator annotated 50 summaries (25 unique and 25 common among all evaluators to compute Inter-Rater Reliability). Results of the evaluation can be found in Table \ref{tab:annotation_quality}. We first conducted a pilot study for evaluation with randomly sampled 10 summaries before proceeding to the final annotation. Table \ref{Table: kappa_appendix} shows the results of Fleiss' Kappa computed on different criteria.
\begin{table}[h]
    \centering
    \resizebox{\columnwidth}{!}{%
    \begin{tabular}{lcccc}
    \toprule
     & \textbf{Human-annotated} & \textbf{GPT-R} & \textbf{GPT-RDQ}\\
    \midrule
    Informativeness & $0.22$ & $0.43$ & $0.45$\\
    Factuality & $0.24$ & $0.36$ & $0.44$\\
    Coherence & $0.25$ & $0.42$ & $0.41$\\
    Conciseness& $0.21$ & $0.38$ & $0.40$\\
    Fluency & $0.24$ & $0.45$ & $0.41$\\
    \midrule
    \textbf{Overall}& $0.23$ & $0.41$ & $0.42$\\
    \bottomrule
    \end{tabular}%
    }
    \caption{\textbf{Fleiss' Kappa.} We compute the Inter-Rater Reliability for human-annotated, GPT-R and GPT-RDQ on five metrics. GPT-R and GPT-RDQ scored higher on all the metrics compared to human summaries.}
    \label{Table: kappa_appendix}
\end{table}%



\section{SDC Approach Effectiveness}
The novelty of our SDC approach lies in utilizing descriptions and question-answer pairs in the selection of pseudo-summaries in the most effective manner. The initial selection based on descriptions and question-answers ensures that the chosen pseudo-summary exhibits information overlap between these sources. This, in turn, aids the model in learning to extract information from these diverse inputs during the summarization process. Moreover, our strategy involves using the selected pseudo-summary to then identify the input reviews that are the most semantically close to it. This dual-step process enhances the model's learning of the opinion summarization task. Table \ref{Table: sdc_appendix} contains the results obtained using different SDC approaches. We find that our approach of creating synthetic datasets performs the best.

\section{MEDOS vs. LLMs?}
Table \ref{Table: llm_summaries} displays a comparison between the summaries generated by the LLM models and our MEDOS model. Our findings reveal that the MEDOS model adeptly captures most user opinions within the summary. However, LLMs go a step further, encompassing additional details to provide a comprehensive perspective on various product aspects. Despite this, our MEDOS model, significantly smaller and reliant solely on unsupervised corpus for synthetic datasets, competently extracts crucial user opinions without the extensive resources and fine-tuning required by LLMs, which often consist of billions of tokens and parameters.

Our primary goal was to leverage existing product data and refine smaller models like BART for multi-source opinion summarization, evaluating their effectiveness compared to ChatGPT. Prioritizing these smaller models aims to enhance accessibility and deployability, particularly on devices with limited resources. While LLMs outshine in performance, our focus on achieving high-quality outputs using smaller models within constraints represents a notable achievement. Insights gained from this endeavor can potentially enhance the data efficiency of larger models in the future. Beyond cost-effectiveness, MEDOS introduces a pathway to substantial results with reduced computational and data needs.

\section{Summary Lengths}
Table \ref{Table: summary_len} reports the mean summary length and mean standard deviations for summaries across three test sets: Amazon, Oposum+, and Flipkart. 
\begin{table}[h]
\centering
\resizebox{\columnwidth}{!}{%
\begin{tabular}{lcccccc}
\toprule
\multicolumn{1}{c}{} & \multicolumn{2}{c}{\textbf{Amazon}} & \multicolumn{2}{c}{\textbf{Oposum+}} & \multicolumn{2}{c}{\textbf{Flipkart}} \\ 
\cmidrule(lr){2-3} \cmidrule(lr){4-5}
\cmidrule(lr){6-7}
 & $\mu$ & $\sigma$ & $\mu$ & $\sigma$ & $\mu$ & $\sigma$ \\
 \midrule
Human annotated & $55.20$ & $12.98$ & $82.16$ & $20.54$ & $118.86$ & $37.11$ \\ 
GPT-R & $58.31$ & $13.01$ & $89.61$ & $8.90$ & $82.71$ & $13.54$ \\ 
GPT-RDQ & $53.64$ & $12.28$ & $81.57$ & $12.88$ & $84.44$ & $12.15$ \\ 
MultimodalSum & $49.03$ & $4.63$ & $46.00$ & $5.33$ & $42.30$ & $4.76$ \\ 
MEDOS & $47.75$ & $5.73$ & $57.36$ & $7.09$ & $51.79$ & $8.28$ \\ 
\bottomrule
\end{tabular}}
\caption{Mean summary length ($\mu$) and mean standard deviation ($\sigma$) for summaries corresponding to the three test sets: Amazon, Oposum+, and Flipkart.}
\label{Table: summary_len}
\end{table}

\section{Example}
Table \ref{Table: example_input} shows the \textit{reviews, product description,} and \textit{question-answers} for a sample product from the Amazon test set. Table \ref{Table: example_summaries_appendix} contains the human-annotated summaries from the original test set and our ChatGPT-generated (GPT-R and GPT-RDQ) summaries followed by different model-generated summaries for the same product.

\clearpage
\begin{table*}[t]
    \centering
    \resizebox{2\columnwidth}{!}{%
    \begin{tabular}{p{2\columnwidth}}
    \toprule
    \small{\textbf{Reviews}}\\
    \midrule
    \small{Exactly as described, at 8 + oz. of solid metal this grip offers a stable way to hold your lightweight digital camera, without putting you fingers in front of the lens or flash. I find it works well with the Kodak PlaySmart Video camera and the Nikon S9100 point and shoot. Opteka HG-5 Pistol Handgrip Stabilizer for Point-n-Shoot, DSLR and Video Cameras}\\
    \midrule
    \small{Probably the best and the least expensive stick I've ever owned and I love it. I use this with my GoPro HD Hero2. It's a bit heavy but the construct is very good. You can use this as a weapon too. lol}\\
    \midrule
    \small{Bought this as part of the stabilizer rig then realized that this was easier to use alone than the rig itself. I am going to use it with a camera with an active stabilizer. Videos looked good. Will update after I use it this weekend. It looks good and is built solid.}\\
    \midrule
    \small{I use this with a dual-camera mount and I like this because of the heft / weight and it stays pretty secure whether I use it with the mount or on my flip video camera or snapshot. I'd recommend this handle highly.}\\
    \midrule
    \small{I was planning to use this with my D7000 + Battery Grip + 80-200 f / 2.8 lens, but when I received it, I changed my mind. It just does not look like it can handle that load. I put it on my Panasonic GF2, and it performs very nicely. Would highly recommend it for lighter cameras.}\\
    \midrule
    \small{The unit is quite sturdy. I bought it to replace the pistol grip unit also featured because the pistol grip locking mechanism did seem to want to lock tight. This unit locks in very tightly and also feels professional. A great purchase for the money.}\\
    \midrule
    \small{This is the 2nd stabilizer that I've purchased one for my Sony a99 and one for my Sony a33. I can't speak highly enough about this handy little item! It's perfectly sized and the ergonomics is ideal! Two thumbs up!!}\\
    \midrule
    \small{A low cost device that I bought and paired with a cell phone reduce jittery videos. Works pretty well for handheld use even when walking. The thread seemed a little recessed at first until I moved the washer flat. I recommend this product for anyone who records videos often for friends, and family especially with your cell phone.}\\
    \midrule
    \small{\textbf{Product Description}}\\
    \midrule
    \small{Opteka HG-1 Heavy-Duty Aluminum Ultra HandGrip Handheld Stabilization System for DSLR and Video Cameras. The Opteka HG-1 HandGrip Stabilization System is a video stabilization device designed specifically for point-and-shoots, Digital SLR cameras and compact camcorders. The Handgrip keeps your hands off the camera and allows you to capture videos from difficult angles. SpecificationsColor:Black; RedMaterials:Aluminum; Foam PaddedThread Size:1/4"Dimensions (HxLxW):6.25" x 1.5" x 1.5" (15.8cm x 3.8cm x 3.8cm)Weight:8.4oz (240g)}\\
    \midrule
    \small{\textbf{Question-Answers}}\\
    \midrule
    \small{What is on the bottom end? Is there a 1/4 - 20 female connection on the bottom?}\\
    \small{Yes it does have a 1/4 - 20 female connection very handy, i hope this helps you.}\\
    \midrule
    \small{Does this work with nikon d800} \\
    \small{It'll work with any camera that has a standard thread tripod socket. Note there is a male post at the top AND a female socket on the bottom. One of the handiest gadgets I've ever bought! If it only came in blue}\\
    \midrule
    \small{Can this be used on a Nikon Coolpix L820, or is that camera too big / insufficient size?} \\
    \small{Yes you can. Use Can be used By any camera or camcorder Threaded for a tripod}\\
    \midrule
    \small{Is the thread 1/4-20} \\
    \small{Yes, 1/4 -20 (1/2 inch long) for standard tripod mount. threads right into the bottom of any small and midsized camera}\\
    \midrule
    \small{If my arm shakes a lot, will this help?} \\
    \small{Probably not. I recommend you check out a mono pod or tripod. There also is a gimble style stabilizer that may help you but I've never used one so if you try it let me know how it works. Hope I answered your question.}\\
    \midrule
    \small{I assume this can be screwed directly into a Canon VIXIA HF20?} \\
    \small{If your camera is threaded for a tripod it will work. It's a little heavy but it's handy to have}\\
    \midrule
    \small{Does this handle water resistant and/or underwater?} \\
    \small{It's just a chunk of machined aluminum with a foam handle. Get it wet, it'll dry off. If it's screwed into an underwater housing you're golden, but if you just dunk your camera under water while attached to this attractive red and black handle, you'll have a nice expensive paperweight Comprende?}\\

    \bottomrule
    \end{tabular}%
    }
    \caption{Reviews, Product Description, and Question-Answers for a sample product from the Amazon test set.}
    \label{Table: example_input}
\end{table*}%

\clearpage
\begin{table*}[t]
    \centering
    \resizebox{2\columnwidth}{!}{%
    \begin{tabular}{p{2\columnwidth}}
    \toprule
    \small{\textbf{Human-annotated summaries}}\\
    \midrule
    \small{Exactly as described. Looks good with a solid build and the unit is quite sturdy. This unit locks in very tightly and also feels professional. It is a great purchase for the money. I'd recommend this handle highly.}\\
    \midrule
    \small{This stabilizer is perfect for light weight cameras and can also be used with a cell phone to help diminish jitter. You can feel the heft in this stabilizer and know that it will stay secure. It feels weighty and professional. I would say to use caution when using it with larger cameras but it is perfect for smaller cameras. Overall, it is recommended.}\\
    \midrule
    \small{A great all-around, sturdy stick for cameras. It is very versatile and quite ideal for lightweight cameras especially, and this stick is highly recommended for its great price-point and overall versatility.}\\
    \midrule
    \small{\textbf{GPT-R summaries}}\\
    \midrule
    \small{The Opteka HG-5 Pistol Handgrip Stabilizer delivers on its promise. With a solid metal build, it secures lightweight cameras effectively. It impressed users across various devices, from Nikon to GoPro, proving its versatility. Sturdy and easy to handle, it's a budget-friendly choice for smoother videos. Whether used independently or as part of a rig, it offers stability and control. An ideal solution for steady shots, even when on the move.}\\
    \midrule
    \small{The Opteka HG-5 Pistol Handgrip Stabilizer is a solid metal grip offering stability for lightweight cameras. Versatile with various camera types, it's praised for its heft, secure hold, and professional feel. Some found it surprisingly effective for reducing jittery videos, while others appreciated its durability and ease of use. Though not suitable for heavy setups, it's ideal for lighter cameras, making it a recommended, cost-effective choice for steady and smooth video capture.}\\
    \midrule
    \small{Opteka HG-5 Pistol Handgrip Stabilizer is a solid metal grip offering stability for lightweight digital cameras like GoPro and point-and-shoots. It's versatile, working well both with rigs and standalone. Sturdy and secure, it handles various cameras effectively. A worthwhile purchase for smoother videos, whether shooting casually or on-the-go.}\\
        \midrule
    \small{\textbf{GPT-RDQ summaries}}\\
    \midrule
    \small{The Opteka HG-1 HandGrip Stabilization System is a versatile tool for capturing steady videos. Its sturdy build provides stability for lightweight digital cameras and camcorders. Users appreciate its ergonomic design and efficient angle adjustments. The grip's thread and weight offer ease of use, enhancing video quality even while on the move. An affordable solution for jitter-free recordings, especially useful for handheld shots and tricky angles.}\\
    \midrule
    \small{Opteka HG-1 HandGrip offers solid stabilization for cameras. With its durable build and foam-padded handle, it adds stability to your shots. It accommodates various cameras, making it versatile for point-and-shoots, DSLRs, and compact camcorders. It\'s especially handy for reducing jittery videos and capturing challenging angles. Its 1/4"-20 thread size ensures compatibility with most devices. The grip is praised for its heft and comfortable ergonomics. A cost-effective tool to improve video stability and angles for diverse cameras.}\\
    \midrule
    \small{The Opteka HG-1 HandGrip Stabilization System proves its worth. Sturdy, versatile, and well-crafted, it stabilizes cameras for smooth video capture. Its ergonomic design offers a secure hold while preventing shaky shots. A valuable addition, especially for lightweight point-and-shoots and smaller cameras. Worth every penny for steady, shake-free videos.}\\
    \midrule
    \small{\textbf{PlanSum}}\\
    \midrule
    \small{i bought this for my nikon d7000 and it works great. i have had no problems with it. the only thing i don't like is that it is a bit bulky, but it's not a big deal for me. if you are looking for a good camera, this is the best way to go.}\\
    \midrule
    \small{\textbf{MultimodalSum}}\\
    \midrule
    \small{This is a great product. I use it with my Canon T3i and it works great. It is a little heavy, but that is to be expected. I like the fact that it has a solid feel to it. I would recommend this product to anyone.}\\
    
    \midrule
    \small{\textbf{BART-concat}}\\
    \midrule
    \small{I bought this to use with the Opteka HG-5 Pistol Handgrip Stabilizer for my Canon Powershot T3i and it works great. It is a bit heavy, but that is to be expected with a tripod.}\\
    \midrule
    \small{\textbf{Our Model (MEDOS)}}\\
    \midrule
    \small{The Opteka HG-1 Heavy-Duty Aluminum Ultra HandGrip Handheld Stabilization System for DSLR and Video Cameras is a great product. I use it with my Nikon Coolpix L820 and it works great. It is a little heavy, but that is to be expected for a small camera.}\\
    \bottomrule
    \end{tabular}%
    }
    \caption{Three human-annotated summaries, three GPT-R summaries, there GPT-RDQ summaries followed by the summaries from different models for the same product as in Table \ref{Table: example_input}.}
    \label{Table: example_summaries_appendix}
\end{table*}%

\clearpage
\begin{table*}[t]
    \centering
    \resizebox{2\columnwidth}{!}{%
    \begin{tabular}{p{2\columnwidth}}
    \toprule
    \small{\textbf{Gold (Amazon GPT-RDQ)}}\\
    \midrule
    \small{The VuPoint Film and Slide Scanner presents a mixed picture. It claims ease \& efficiency for converting film to digital format, but user experiences differ. While some commend its user-friendliness and recommend it, others report frustrating issues like bleached images, compatibility problems, and subpar results. The need for XP compatibility limits its usefulness for modern systems. It offers potential for simple scanning but may require persistence to achieve desired outcomes.}\\
    \midrule
    \small{\textbf{Chatglm2-6b}}\\
    \midrule
    \small{scanner from VuPoint. The scanner has worked great for me and I have been able to scan over 1, 200 35mm negatives and a dozen slides. I have also been able to upload a few black and white negatives from 1963. The scanner is user-friendly and efficient tool. The software that comes with the scanner is good for Windows XP and Vista, but the company prefers to sell other products. I am looking for a scanner that can work with more modern versions of Windows, but I am not satisfied with the VuPoint company. I am also waiting for a replacement unit to be delivered.}\\
    \midrule
    \small{\textbf{Claude-2}}\\
    \midrule
    \small{Reviews of this scanner are quite mixed. Some find it easy to use and good for digitizing old negatives and slides. But many cite issues getting it to work properly, with several noting it is not compatible with Windows versions after XP. Complaints include poor image quality, too much light bleaching images out, and tech support being expensive. The included software also gets some negative feedback for lacking updates. A few reviewers say the holders are fiddly for fitting negatives into. Overall while some find it efficient for converting film, many note disappointing quality and outdated software/hardware compatibility as negatives.}\\
    \midrule
    \small{\textbf{Llama-2-7b-chat}}\\
    \midrule
    \small{The VuPoint FS-C1-VP Film and Slide Digital Converter has received mixed reviews. Some users have had issues with image quality, light bleaching, and poor customer support, while others have found the scanner easy to use and effective at converting slides and negatives to digital format. The product's compatibility with newer operating systems and software updates has also been a concern for some reviewers. Overall, the product's performance and customer service have been a disappointment for some buyers.}\\
        \midrule
    \small{\textbf{Llama-2-70b-chat}}\\
    \midrule
    \small{The VuPoint FS-C1-VP Film and Slide Digital Converter has mixed reviews. Some reviewers find it easy to use and efficient for scanning 35mm film negatives and slides, while others have issues with image quality, software compatibility, and customer support. One reviewer had a positive experience with the company's customer service, while another found it unhelpful. The product's ability to scan images at 5.0 Mega pixel quality and automatic color balance and exposure control are mentioned in the description, but some reviewers still express dissatisfaction with the product's overall performance.}\\
    
    \midrule
    \small{\textbf{Our Model (MEDOS)}}\\
    \midrule
    \small{I purchased the VuPoint FS-C1-VP Film and Slide Digital Converter to scan my 35mm film and slide negatives. It is not compatible with Windows XP. The software does not work with Windows 7 or 8. I have tried to contact the company and they do not respond to my emails. I would not recommend this product to anyone.}\\
    \bottomrule
    \end{tabular}%
    }
    \caption{\textbf{Comparative analysis with LLM generated summaries.} ChatGPT-generated summary using \textit{reviews, description, and question-answers} (GPT-RDQ) followed by different LLM-generated summaries and our MEDOS model generated-summary for an Amazon test set product.}
    \label{Table: llm_summaries}
\end{table*}%

\end{document}